\documentclass[10pt,twocolumn,letterpaper]{article}

\usepackage[pagenumbers]{cvpr} 

\usepackage{graphicx}
\usepackage{amsmath}
\usepackage{amssymb}
\usepackage{booktabs}
\usepackage{bm}
\usepackage{physunits}

\usepackage{cuted}
\usepackage{capt-of}

\usepackage{multirow}
\usepackage{verbatim}
\usepackage{comment}
\usepackage{adjustbox}
\usepackage{wrapfig}

\usepackage{url}            
\usepackage{amsfonts}       
\usepackage{nicefrac}       
\usepackage{microtype}      

\usepackage{array}
\usepackage{xspace}
\usepackage{pifont}

\usepackage[dvipsnames]{xcolor}
\usepackage{colortbl}

\usepackage{tabularx}

\newcommand{\myparagraph}[1]{\vspace{0.1em}\noindent\textbf{#1}}

\definecolor{cvprblue}{rgb}{0.21,0.49,0.74}
\usepackage[pagebackref,breaklinks,colorlinks,allcolors=cvprblue]{hyperref}

\def\paperID{2061} 
\def\confName{CVPR}
\def\confYear{2025}

\title{SMGDiff: Soccer Motion Generation using diffusion probabilistic models}

\author{
    Hongdi Yang \textsuperscript{\rm 1*}\quad
    Chengyang Li \textsuperscript{\rm 1*}\quad
    Zhenxuan Wu \textsuperscript{\rm 1}\quad
    Gaozheng Li \textsuperscript{\rm 1}\quad
    Jingya Wang \textsuperscript{\rm 1}\\
    Jingyi Yu \textsuperscript{\rm 1}\quad
    Zhuo Su \textsuperscript{\rm 2}\quad
    Lan Xu \textsuperscript{\rm 1\dag}\\
    \textsuperscript{\rm 1} ShanghaiTech University  \quad
    \textsuperscript{\rm 2} ByteDance\\
    \
}

\begin{document}
\maketitle
\begin{strip}
    \centering
    \vspace{-40px}
    \captionsetup{type=figure}
    \includegraphics[width=\textwidth]{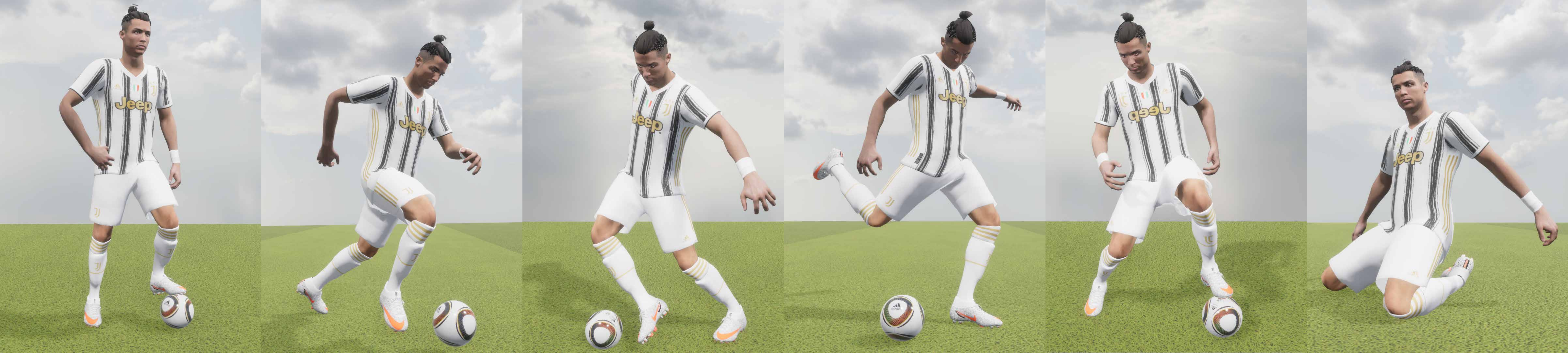}
    \caption{Our method, SMGDiff, enables users to control soccer motions based on character displacement and soccer skill, simulating an interactive gameplay experience. It can generate a diverse range of high-quality soccer motions while ensuring real-time performance.} 
    \label{fig:teaser}
\end{strip}

\begin{abstract}
Soccer is a globally renowned sport with significant applications in video games and VR/AR. However, generating realistic soccer motions remains challenging due to the intricate interactions between the human player and the ball. In this paper, we introduce SMGDiff, a novel two-stage framework for generating real-time and user-controllable soccer motions. Our key idea is to integrate real-time character control with a powerful diffusion-based generative model, ensuring high-quality and diverse output motion. In the first stage, we instantly transform coarse user controls into diverse global trajectories of the character. In the second stage, we employ a transformer-based autoregressive diffusion model to generate soccer motions based on trajectory conditioning. We further incorporate a contact guidance module during inference to optimize the contact details for realistic ball-foot interactions. Moreover, we contribute a large-scale soccer motion dataset consisting of over 1.08 million frames of diverse soccer motions. Extensive experiments demonstrate that our SMGDiff significantly outperforms existing methods in terms of motion quality and condition alignment. 

\end{abstract}    
\section{Introduction} 

\label{sec:intro}
Soccer is widely recognized as the most popular sport globally, leading to a growing emphasis on generating and controlling soccer motions in various games and VR/AR applications. However, real-time generation of realistic soccer motion in a highly controllable manner remains challenging. This challenge primarily stems from the complex interactions between the player and the ball, especially for the intricate human trajectories and the precise ball-foot contacts. 
Soccer motion has been extensively studied in both industry and academia. 
Popular soccer games like \textit{EA SPORTS FC} series~\cite{eaSportsFc} feature impressive soccer animations with interactive and real-time character control. Yet, they heavily rely on example-based motion matching techniques, leading to visual artifacts.
In academia, various approaches~\cite{10.1145/3306346.3322963,liu2021motorcontrolteamplay,peng2021amp,peng2019mcplearningcomposablehierarchical} focus on enhancing the physical realism of soccer motion control using reinforcement learning or physics-based simulation. Yet, these methods are restricted to specific skills of soccer motion. For example, Hong et al.~\cite{10.1145/3306346.3322963} performed physical-based control for dribbling and shooting, while Xie et al.~\cite{2022-Soccer-Juggle} utilized a layer-wise structure only for soccer juggling. 
Nevertheless, a high-quality soccer motion generation scheme should cover a diverse spectrum of soccer skills, from dribbling to playing tricks, from shooting to celebrating.

Recent advancements in motion diffusion probabilistic models~\cite{guo2022generating,zhang2022motiondiffuse,tevet2023humanmotion,tseng2022edge,Liang_2024_CVPR} have brought significant potential for capturing complex data distribution and generating diverse motions. Some very recent variants~\cite{zhou2023emdm,Zhao:DART:2024,camdm} further accelerate the diffusion models for efficient on-the-fly motion generation.
Notably, CAMDM~\cite{camdm} utilizes efficient autoregressive diffusion to generate diverse character animation, responding to user-supplied control signals. Yet, these methods focus solely on human locomotion with diverse style transitions, instead of the soccer motions with complex human-object interactions. Additionally, some diffusion-based methods~\cite{CG-HOI,ghosh2023imos,li2023task,tevet2024closd} generate realistic human-object interactions but require time-consuming post-optimization, unsuitable for real-time interactable applications.
In a nutshell, it lacks a viable solution for real-time user-controllable soccer motion generation. The challenges stem from the limited availability of soccer motion datasets and the absence of an effective generation framework tailored for soccer motion.

In this paper, we present \textit{SMGDiff}, a motion diffusion model for real-time and user-controllable soccer motion generation. As shown in Fig.~\ref{fig:teaser}, given diverse user-supplied control signals like desired character speed, direction, and soccer skills, SMGDiff can generate high-quality and natural soccer motions in real-time, which was unseen before.
Specifically, we first collect a large-scale soccer motion dataset, named \textit{Soccer-X} for our generative model. In the Soccer-X dataset, we utilize 16 OptiTrack cameras~\cite{Optitrack} to reconstruct ground-truth 3D human and soccer motions. It covers six common soccer motion categories, comprising over 1.08 million frames and more than 10 hours of diverse and high-quality motion data from 30 soccer players. To the best of our knowledge, our Soccer-X dataset is the first of its kind for data-driven soccer motion generation featuring a wide range of soccer skills.

With our unique dataset, we further propose an efficient two-stage framework for controllable soccer motion generation. 
\textbf{In the first stage}, we generate the global trajectory of the character (both the facing direction and location) from the coarse user-supplied control signals. Such a global trajectory serves as a strong conditioning for the subsequent soccer motion diffusion model. We adopt a single-step diffusion model based on a transformer encoder, for instant and diverse trajectory generation. We further utilize a temporal blending strategy to ensure the smoothness of the generated trajectory, responding in real-time to diverse control signals.
\textbf{In the second stage}, we tailor a transformer-based autoregressive diffusion model~\cite{camdm} for efficient soccer motion generation, based on the trajectory conditioning from the first stage. 
Specifically, we jointly consider the skeletal human motion, rigid ball motion, and contact information in this soccer motion diffusion model. 
We further introduce a contact guidance module to refine ball-foot interactions during the inference process of the diffusion model.
In our contact guidance module, when the moving speed of the ball significantly changes and ball-foot contact happens, a novel contact loss function is activated to adjust the sampling result of soccer motions.

We conduct a series of experiments to demonstrate the effectiveness of our SMGDiff approach. We also showcase a real-time demo for vivid soccer motion generation with diverse soccer skills from user controls.
In summary, our contributions are as follows:

\begin{itemize} 
    \setlength\itemsep{0em}
    \item We introduce a novel motion diffusion model for real-time, user-controllable soccer motion generation, achieving state-of-the-art performance.
    
    \item We propose a highly efficient trajectory generation module, as well as a novel contact guidance module to generate diverse and accurate ball-foot contacts.
    
    \item We contribute a large-scale motion dataset that includes high-quality soccer motion annotations. Our data and model will be disseminated to the community.
\end{itemize}

\begin{figure*}[t!]
    \centering
    \includegraphics[width=\textwidth]{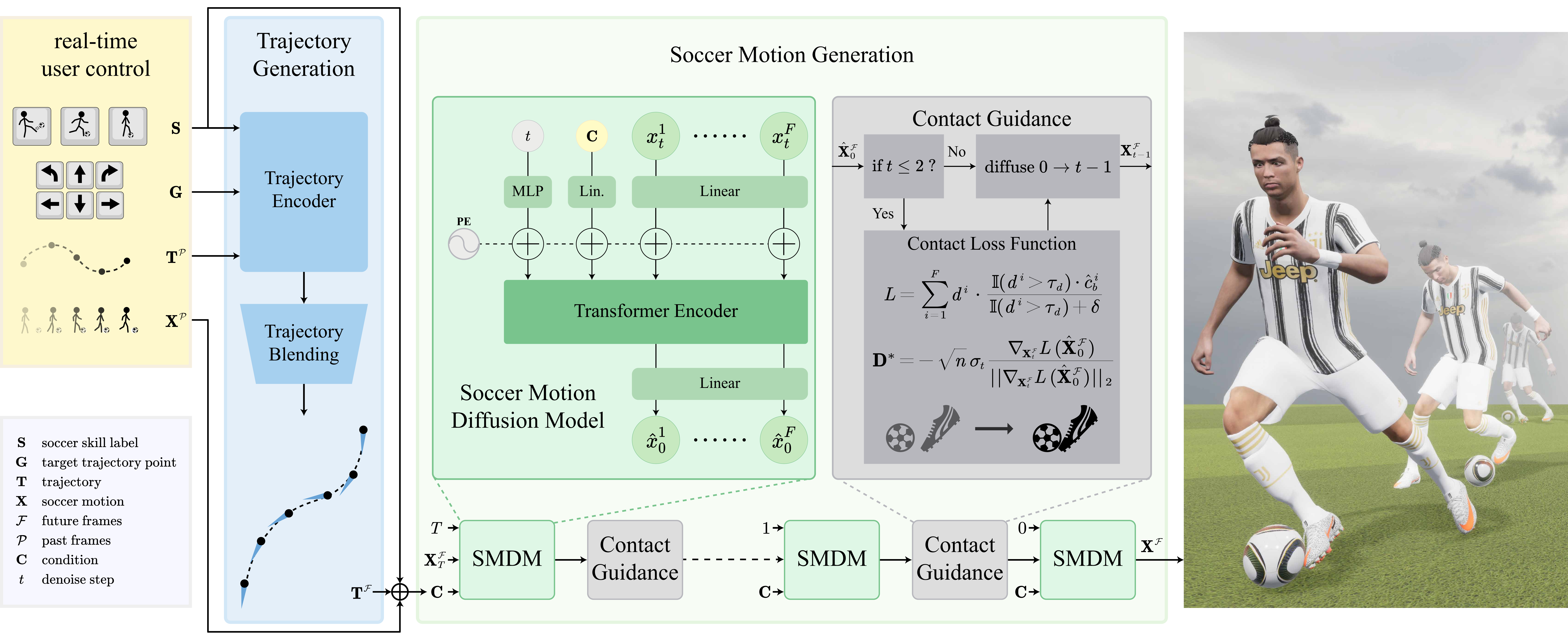}
    \caption{Pipeline of SMGDiff. Our framework consists of two stages: 
    In the trajectory generation stage, we transform soccer skill label $\mathbf{S}$, target trajectory point $\mathbf{G}$ from user control, and past trajectory $\mathbf{T}^{\mathcal{P}}$ into refined future trajectory $\mathbf{T}^{\mathcal{F}}$. 
    In the soccer motion generation stage, the soccer motion diffusion model is fed with a noisy motion sequence $\mathbf{X}_{T}^{\mathcal{F}}$ and condition information $\mathbf{C}$, which concatenates $\mathbf{S}$, $\mathbf{T}^{\mathcal{F}}$ and past motion $\mathbf{X}^{\mathcal{P}}$. Contact guidance module refines the predicted soccer motion $\hat{\mathbf{X}}_{0}^{\mathcal{F}}$ during the diffusion process to enhance the contact details.}
    \label{fig:pipeline}
\end{figure*}
\section{Related Work}
\label{sec:related_work}
\myparagraph{Real-Time Motion Controller.}
Real-time motion control holds significant value in the gaming domain. The earliest approaches were based on motion matching~\cite{clavet2016motion,holden2020learned,mach2021motion,codebookmatching} and state machines~\cite{lau2005behavior,starke2019neural,ArikanForsyth2002,Kovar2002}. 
Later, numerous methods~\cite{fragkiadaki2015recurrent,li2017auto,holden2015learning,holden2016deep,starke2019neural,learnedmm,motionrecom,mason2022real,starke2021neural} have employed neural networks in character animation to enhance motion diversity. Additionally, physics-based approaches~\cite{lee2018interactive,lee2021learning,Hodgins:2017:DOE,ye2022neural3points,ControlVAE,yao2023moconvqunifiedphysicsbasedmotion} further enhanced the realism of character animations. Recent methods~\cite{pfnn,Starke2020,mann} have also incorporated feature extraction techniques. However, these methods, relying on deterministic networks, encounter challenges in generating high-quality and diverse motion sequences.

\myparagraph{Motion Diffusion.}
Probabilistic generative models~\cite{nichol2022glide,saharia2022photorealistic,ho2020denoising} have demonstrated the ability to produce high-quality and diverse motions~\cite{tevet2023humanmotion,zhang2022motiondiffuse,kim2023flame}. A significant body of work incorporates various conditions during generation, including text~\cite{chen2023executing,liang2023intergen,yuan2023physdiff,zhang2022motiondiffuse,Liang_2024_CVPR,kim2023flame}, music~\cite{Alexanderson2022ListenDA,tseng2022edge,Zhou_2023_ude}, action~\cite{zhao_2023_Modiff,Zhao2024DenoisingDP,findlay2022denoising}, speech~\cite{ao2023gesturediffuclip,zhao2024media2face,Chen2023DiffusionTalkerPA}, incomplete motion sequences~\cite{tevet2023humanmotion,xie2023omnicontrol,Zhang_2023_controlnet}, trajectory~\cite{gmd,odn}, and scene~\cite{Li2023ObjectMG,Huang_2023_diffu_based_gen}. Recent methods, such as CAMDM~\cite{camdm} have enabled motion generation conditioned on user-controlled style transitions. Building on prior work, we apply motion diffusion models to soccer, a domain characterized by various complex motions.

\myparagraph{Human-Object Interactions.}
Previous works on Human-Object Interaction (HOI) generation have employed physics-based methods~\cite{xie2023hierarchicalplanningcontrolbox, merel2020catchcarryreusable, hassan2023synthesizingphysicalcharactersceneinteractions} and kinematic-based approaches~\cite{pfnn, mann, 10.1145/3386569.3392450, 10.1145/3355089.3356505}. More recent approaches~\cite{CG-HOI, ghosh2023imos, li2023task} integrate physical constraints during inference to further boost realism. However, these methods often incur significant computational overhead, rendering them impractical for interactive applications like soccer that require low-latency responses. Popular soccer game EA SPORTS FC series~\cite{eaSportsFc} relies on extensive pre-recorded data for motion matching rather than real-time motion synthesis. Prior soccer-related research~\cite{10.1145/3306346.3322963,liu2021motorcontrolteamplay,peng2019mcplearningcomposablehierarchical,Peng_2021, 2022-Soccer-Juggle} has largely concentrated on specific actions, such as shooting, dribbling, and juggling, without covering the broader range of soccer motions. In contrast, we present a large-scale open-source soccer dataset and a novel framework to cover a diverse spectrum of soccer skills.

\section{Method} 
\label{sec:method}
Our objective is to generate realistic and diverse soccer motions while enabling real-time user control. To this end, we propose a two-stage framework consisting of trajectory generation and soccer motion generation, as depicted in Fig.~\ref{fig:pipeline}. 
In the trajectory generation stage (Sec.~\ref{sec:Trajectory Generation Part}), coarse user control signals are transformed into refined trajectories, providing strong conditioning for the motion synthesis stage.
In the soccer motion generation stage, we employ a soccer motion diffusion model (Sec.~\ref{sec:Motion Diffusion Model}) to produce a variety of soccer motions. Additionally, a contact guidance module (Sec.~\ref{sec:Contact Guidance}) is implemented to refine ball-foot contacts.

\myparagraph{Representation.}
For clarity in presentation, we define a set of notations. First, the user selects a soccer skill label $\mathbf{S} \in$ \{dribble, trick, shoot, stand, celebrate, off-the-ball move\} and specifies character direction and speed via keyboard arrow keys and press intensity. We multiply the character speed by the normalized direction vector and then add it to the character's current position to get the target trajectory point $\mathbf{G}$.

Under user control, the trajectory generation model generates the future trajectory $\mathbf{T}^{\mathcal{F}}$, which includes the future human root position and orientation projected onto the ground. Superscripts $\mathcal{F} = \{1, 2, \dots, F\}$ denote future frame indices and $\mathcal{P} = \{P, P+1, \dots, -1\}$ denotes past frame indices, where $P$ and $F$ are the total numbers of past and future frames, respectively.

Finally, the soccer motion generation model takes soccer skill label $\mathbf{S}$, future trajectory $\mathbf{T}^{\mathcal{F}}$ and past soccer motion $\mathbf{X}^{\mathcal{P}}=\{x^{i}\}_{i \in \mathcal{P}}$ as condition $\mathbf{C}$ and generates future soccer motion $\mathbf{X}^{\mathcal{F}} = \{x^{i}\}_{i \in \mathcal{F}}$. In this context, $x^{i} = \{h, b, c\}$ is the representation of soccer motion at frame $i$, comprising three components: human state $h$, ball state $b$, and binary contact label $c$. Human state $h \in \mathbb{R}^{3 + 24 \times 6}$ includes root position $h_{p}$ and joints rotation $h_{\theta}$. We use the 24-joint SMPL format~\cite{SMPL} for the human skeleton and a 6-DOF representation~\cite{R6D} for rotation and direction. Ball state $b \in \mathbb{R}^{3+3+1}$ is composed of relative ball position $b_p'$, global ball velocity $b_v$ and ball control weight $w_b$  with reference to Starke et. al.~\cite{10.1145/3386569.3392450}. Ball control weight $w_b$ is utilized to transform the ball's global position $b_p$ to relative ball position $b_p'$ which is defined as:
\begin{equation}
    w_b = 1 - ||b_{p}^{xy} - h_{p}^{xy}|| / r, 
    \label{eq:ball control weight}
\end{equation}
where $b_p^{xy}$ and $h_p^{xy}$ represent the horizontal global position of the ball and human root, $r=2m$ is the radius around the human root. We obtain relative ball position $b_p'$:
\begin{equation}
    b_p' = w_b \cdot (b_p - h_p).
    \label{eq:ball position}
\end{equation}
The binary contact label $c = \{c_g, c_b\}$ consists of two components: $c_g$ represents the contact between the foot joints and the ground, while $c_b$ indicates the contact between the foot joints and the ball. A value of 1 indicates contact and 0 indicates no contact.


\subsection{Trajectory Generation Model} \label{sec:Trajectory Generation Part}
We employ a single-step diffusion model to transform coarse user control signals into diverse character trajectories, followed by a blending strategy to ensure smoothness.

\myparagraph{Network Structure.}
To achieve real-time performance and efficiency, we deploy a lightweight single-step diffusion model based on a transformer encoder. The network's conditional inputs include soccer skill label $\mathbf{S}$, target trajectory point $\mathbf{G}$, and past trajectory $\mathbf{T}^{\mathcal{P}}$.  Additionally, the input conditions are concatenated with a random Gaussian noise $\epsilon\sim\mathcal{N}(0, \mathbf{I})$ to enhance the trajectory diversity. During runtime, we sample the random noise $\epsilon$ and the model inference would be similar to a single-step diffusion reverse process, as defined in DDPM~\cite{DDPM}:
\begin{equation}
\begin{aligned}
&p_\theta(\mathbf{z}_{0:1}):=\epsilon p_\theta(\mathbf{z}_{0}|\mathbf{z}_1), \\
&p_\theta(\mathbf{z}_{0}|\mathbf{z}_1):=\mathcal{N}(\mathbf{z}_{0};\mathbf{\mu}_\theta(\mathbf{z}_1,1),\mathbf{\Sigma}_\theta(\mathbf{z}_1,1)),
\end{aligned}
\end{equation}

where $\mathbf{z}$ denotes the predicted future trajectory and the total timestep of diffusion is set to 1. During training, we employ a simple reconstruction loss and a velocity loss to optimize the network. For a more detailed explanation of the network structure, please refer to the supplementary materials.

\myparagraph{Trajectory Blending.}
At runtime, we adopt the \textit{Heuristic Future Trajectory Extension (HFTE)} approach from CAMDM~\cite{camdm} to balance motion smoothness and consistency with user control. When the user inputs a new control signal, the newly generated trajectory results from the trajectory generation model will be blended with the previous generation output to prevent abrupt changes in the character’s direction and speed. The blended trajectory $\mathbf{T}^{\mathcal{F}}$ is then used as the conditional input for the subsequent stage.

\subsection{Soccer Motion Diffusion Model} \label{sec:Motion Diffusion Model}
We deploy a transformer-based autoregressive diffusion model to generate high-quality and diverse soccer motions based on trajectory conditioning. 

\myparagraph{Network Structure.}
Our method utilizes the diffusion probabilistic model to synthesize soccer motion. Following the definition of diffusion in DDPM~\cite{DDPM}, the forward diffusion process is formulated as a Markov chain:
\begin{equation}
    q(\mathbf{X}_{t}^{\mathcal{F}} | \mathbf{X}_{0}^{\mathcal{F}}) = \mathcal{N}(\sqrt{\bar{\alpha}_{t}} \mathbf{X}_{0}^{\mathcal{F}}, (1-\bar{\alpha}_{t}) \mathbf{I}),
\end{equation}
where $\mathbf{X}^{\mathcal{F}}_{0}$ is the future soccer motion drawn from data distribution, $\bar{\alpha}_{t} \in (0,1)$ are constants hyper-parameters, when $\bar{\alpha}_{t}$ approaches 0, we can approximate $\mathbf{X}_{t}^{\mathcal{F}} \sim \mathcal{N}(0, \mathbf{I})$.

In the reverse process, we follow Ramesh et al.~\cite{Ramesh2022HierarchicalTI} to directly predict $\mathbf{X}_{0}^{\mathcal{F}}$ itself as $\hat{\mathbf{X}}_{\phi}^{\mathcal{F}}(\mathbf{x}_{t}, t, \mathbf{C})$, where $\phi$ represents the parameters of the neural network. $\mathbf{C} = \{ \mathbf{S}, \mathbf{X}^{\mathcal{P}}, \mathbf{T}^{\mathcal{F}} \}$ is the formulated condition information, including soccer skill label $\mathbf{S}$, past soccer motion $ \mathbf{X}^{\mathcal{P}}$ and future trajectory $\mathbf{T}^{\mathcal{F}}$. We optimize $\phi$ with the \textit{simple} objective introduced in Go et al.~\cite{DDPM}:
\begin{equation}
    \mathcal{L}_{\text{simple}} = \mathbb{E}_{\mathbf{X}_{0}^{\mathcal{F}}, t} [ ||\mathbf{X}_{0}^{\mathcal{F}} - \hat{\mathbf{X}}^{\mathcal{F}}_{\phi}(\mathbf{X}_{t}^{\mathcal{F}}, t, \mathbf{C})||_{2}^{2} ].
    \label{eq:simple loss}
\end{equation}
We also incorporate auxiliary losses referenced from Tevet et al.~\cite{tevet2023humanmotion} to promote three aspects of physical realism: joint position (Eq.~\ref{eq:pos loss}), velocities (Eq.~\ref{eq:vel loss}) and foot contact (Eq.~\ref{eq:contact loss}).
\begin{equation}
    \mathcal{L}_{\text{pos}} = \frac{1}{F} \sum_{i=1}^{F} ||FK(x_{0}^{i}) - FK(\hat{x}_{0}^{i})||_{2}^{2},
    \label{eq:pos loss}    
\end{equation}
\begin{equation}
    \mathcal{L}_{\text{vel}} = \frac{1}{F-1} \sum_{i=1}^{F-1} ||(x_{0}^{i+1} - x_{0}^{i}) - (\hat{x}_{0}^{i+1} - \hat{x}_{0}^{i})||_{2}^{2},
    \label{eq:vel loss}
\end{equation}
\begin{equation}
    \mathcal{L}_{\text{foot}} = \frac{1}{F-1} \sum_{i=1}^{F-1} 
    ||(FK(\hat{x}^{i+1}_{0}) - FK(\hat{x}^{i}_{0})) \cdot c_{g}^{i} ||_{2}^{2},
    \label{eq:contact loss}
\end{equation}
where $FK(\cdot)$ denotes the forward kinematic function converting human joint rotations into human joint global positions (only apply to foot joints in Eq.~\ref{eq:contact loss}), $c_{g}^{i}$ is the binary contact mask between foot and ground for each frame $i$. 
Our final training loss is:
\begin{equation}
    \mathcal{L} = \mathcal{L}_{\text{simple}} + \lambda_{\text{pos}}\mathcal{L}_{\text{pos}} + 
    \lambda_{\text{vel}}\mathcal{L}_{\text{vel}} + 
    \lambda_{\text{foot}}\mathcal{L}_{\text{foot}}.
\end{equation}
For more details on the soccer motion diffusion model, please refer to the supplementary materials.
\subsection{Contact Guidance} \label{sec:Contact Guidance}
During inference, we integrate a contact guidance module that utilizes an off-the-shelf loss guidance mechanism within the diffusion process to improve contact fidelity. Specifically, we develop a specialized contact loss function to refine the ball-foot contacts.

\myparagraph{Contact Loss Function.}
After the soccer motion generation model generates the future soccer motion sample $\hat{\mathbf{X}}_{0}^{\mathcal{F}}$, we get the future ball position $b_p$ by applying the inverse transformation of Eq.~\ref{eq:ball position}. 
We can also get ball acceleration $b_a$ which is obtained by applying a second-order difference to the ball's global positions. 

According to the laws of mechanics, the ball undergoes frictional forces from the ground, resulting in a small, constant acceleration. When the ball's acceleration exceeds a certain threshold, we infer that the ball is influenced by both ground friction and the force exerted by the foot. In this case, we confirm that contact exists between the ball and foot joints, denoted as $\hat{c}_b$.
\begin{equation}
    \hat{c}_b = \mathbb{I} (||b_a|| > \tau_{a}), 
\end{equation}
where $\mathbb{I}$ is the indicator function, $\tau_{a} = 2 m/s^{2}$ signifies the acceleration threshold. 
Next, we identify the appropriate foot joint for ball contact. Based on real-life soccer motion, the lifted foot joint is given priority. We compute $d$, the distance between the ball and each potential contact joint in every frame:
\begin{equation}
   d = \min_{j \in \text{foot joints}} ((f_{p}^{j} - b_{p}) \cdot (1 + (w_{d} - 1) \cdot c_{g}^{j})), 
\end{equation}
where superscript $j$ denotes the joint index, $f_{p}^{j}$ is the global position of the joint, $c_g^j$ indicates whether the joint is in contact with the ground, $w_{d} = 2$ is a penalty term that encourages the lifted foot to make contact with the ball.
Finally, our contact loss function for guidance is
\begin{equation}
    L = \sum_{i = 1}^{F} d^{i} \cdot \frac{\mathbb{I}(d^{i} > \tau_{d}) \cdot \hat{c}_b^{i}}{\mathbb{I}(d^{i} > \tau_{d}) + \delta},
\end{equation}
where superscript $i$ denotes the frame index, $\tau_{d} = 0.1m$ is the contact distance threshold and $\delta$ is a small constant to prevent division by zero. Our contact loss function precisely identifies the foot joint designated to make contact with the ball and incorporates the distance metric as the loss value, effectively guiding the contact between the foot and the ball.

We adopt the loss guidance approach from DSG~\cite{yang2023dsg}, which proposes larger, adaptive step sizes while preserving the original data distribution. Initially, we compute the steepest gradient descent direction: 
\begin{equation}
    \mathbf{D}^{*} = - \sqrt{n} \sigma_{t} \frac{\nabla_{\mathbf{X}_{t}^{\mathcal{F}}} L(\hat{\mathbf{X}}_{0}^{\mathcal{F}})}{||\nabla_{\mathbf{X}_{t}^{\mathcal{F}}} L(\hat{\mathbf{X}}_{0}^{\mathcal{F}})||_{2}},
\end{equation}
where $n$ represents the data dimensions and $\sigma_{t}$ is the standard deviation in the diffusion process. 
Subsequently, we weigh the gradient direction and the unconditional sampling direction to enhance diversity:
\begin{equation}
    \mathbf{D} = \mathbf{D}_{t} + w_{r} (\mathbf{D}^{*} - \mathbf{D}_{t}),
\end{equation}
\begin{equation}
    \mathbf{X}_{t-1}^{\mathcal{F}} = \mu_{t} + \sqrt{n} \sigma_{t} \mathbf{D} / ||\mathbf{D}||,
\end{equation}
where $\mathbf{D}_{t} = \sigma_{t} \epsilon_{t} (\epsilon_{t} \sim \mathcal{N}(0,\mathbf{I}))$ represents the unconditional sampling direction, $\mathbf{D}$ is the weighted direction, $w_{r} = 0.5$ is the guidance rate and $\mu_{t}$ is the mean prediction.

\subsection{Implementation Details}

\myparagraph{Network Parameters.}
To enable real-time soccer motion generation with user control, we meticulously calibrate our system parameters. During both training and inference, we use a 30Hz frame rate for the generated motions, with past frame number $P=10$ and future frame number $F=45$ to facilitate short-sequence predictions. Additionally, we configure the denoise step in the diffusion model to 8, ensuring real-time inference speed.

\myparagraph{Employment Strategy.}
Due to the time-consuming nature of gradient calculations in contact guidance, it’s impractical to perform contact guidance at every denoise step. We argue that applying contact guidance is unnecessary when diffusion noise is high, as the predicted motion is not sufficiently refined, and incorrect guidance may lead to unnatural results. Hence, we restrict the application of contact guidance to the last two steps of the diffusion model. Systematic experiments of the employment strategy are presented in Sec.~\ref{sec:live demo}.

\myparagraph{Runtime Environment.}
For the environmental setup, we implement real-time character control through network communication between the Unity engine and Python. Initially, Unity captures user keyboard inputs and generates future trajectories using the employed trajectory generation model. The trajectory conditioning is then transmitted to Python via a TCP network. Utilizing the soccer motion generation model, the Python side generates human translation, joint rotations, and ball translation. The outputs are subsequently sent back to Unity for visualization, with the ball’s rotation computed using Unity’s physics engine. All model inferences and network communications are executed on a single machine equipped with an Intel Core i7-10700K CPU and an NVIDIA GeForce RTX 3080 Ti GPU.
\begin{figure}[t!]
    \centering
    \includegraphics[width=\linewidth]{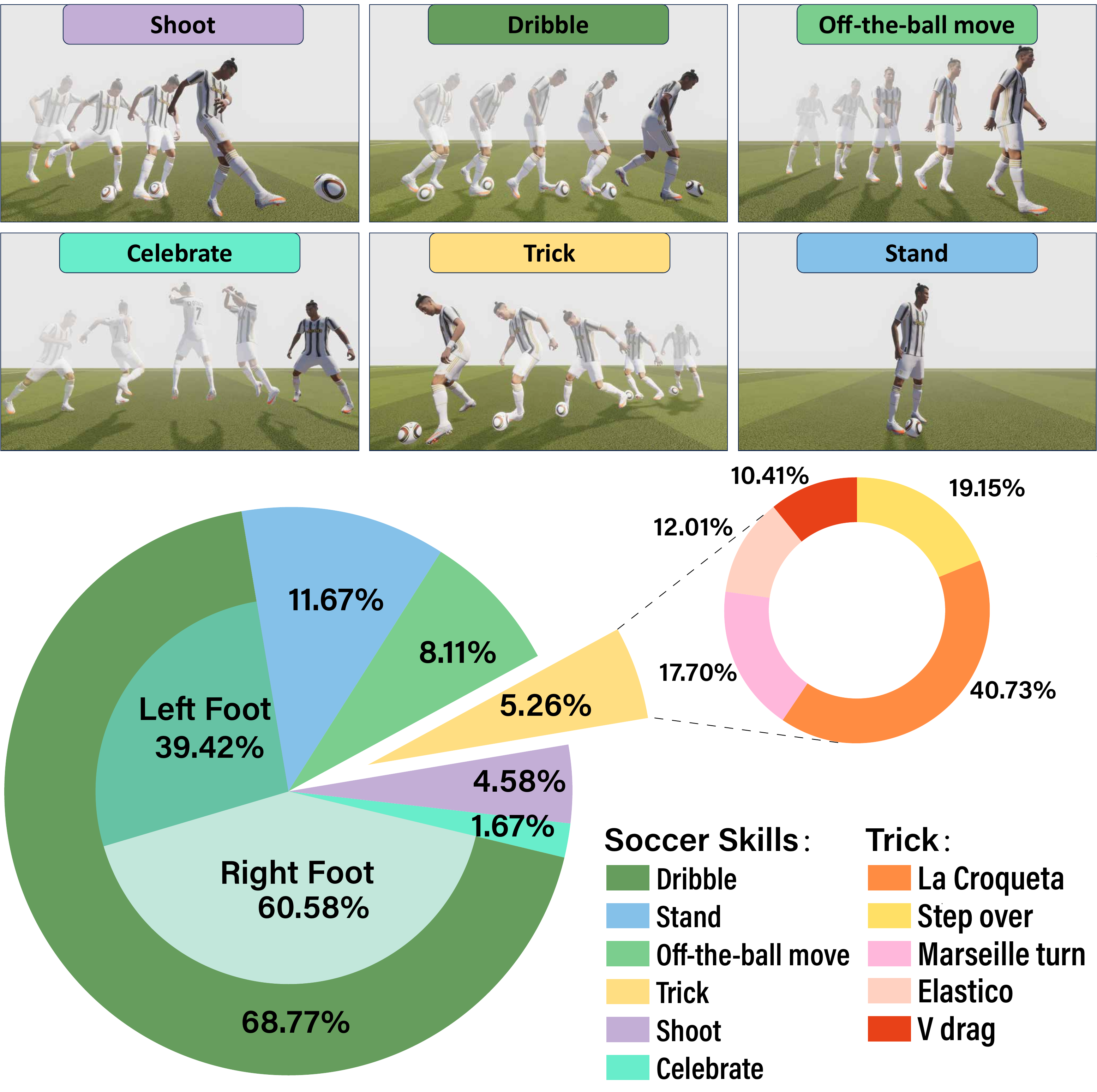}
    \caption{The top section exhibits selected highlights of our dataset. The bottom section features a proportion of different soccer motions. In total, our dataset comprises 2398 sequences and captures approximately 1.08 million frames of data.}
    \label{fig:dataset}
\end{figure}

\section{Dataset}
\label{sec:dataset}
To train and evaluate model performance, we construct \textit{Soccer-X}, a high-quality human-soccer interaction dataset covering diverse soccer motions.

\begin{figure*}[t]
    \centering
    \includegraphics[width=1.0\textwidth]{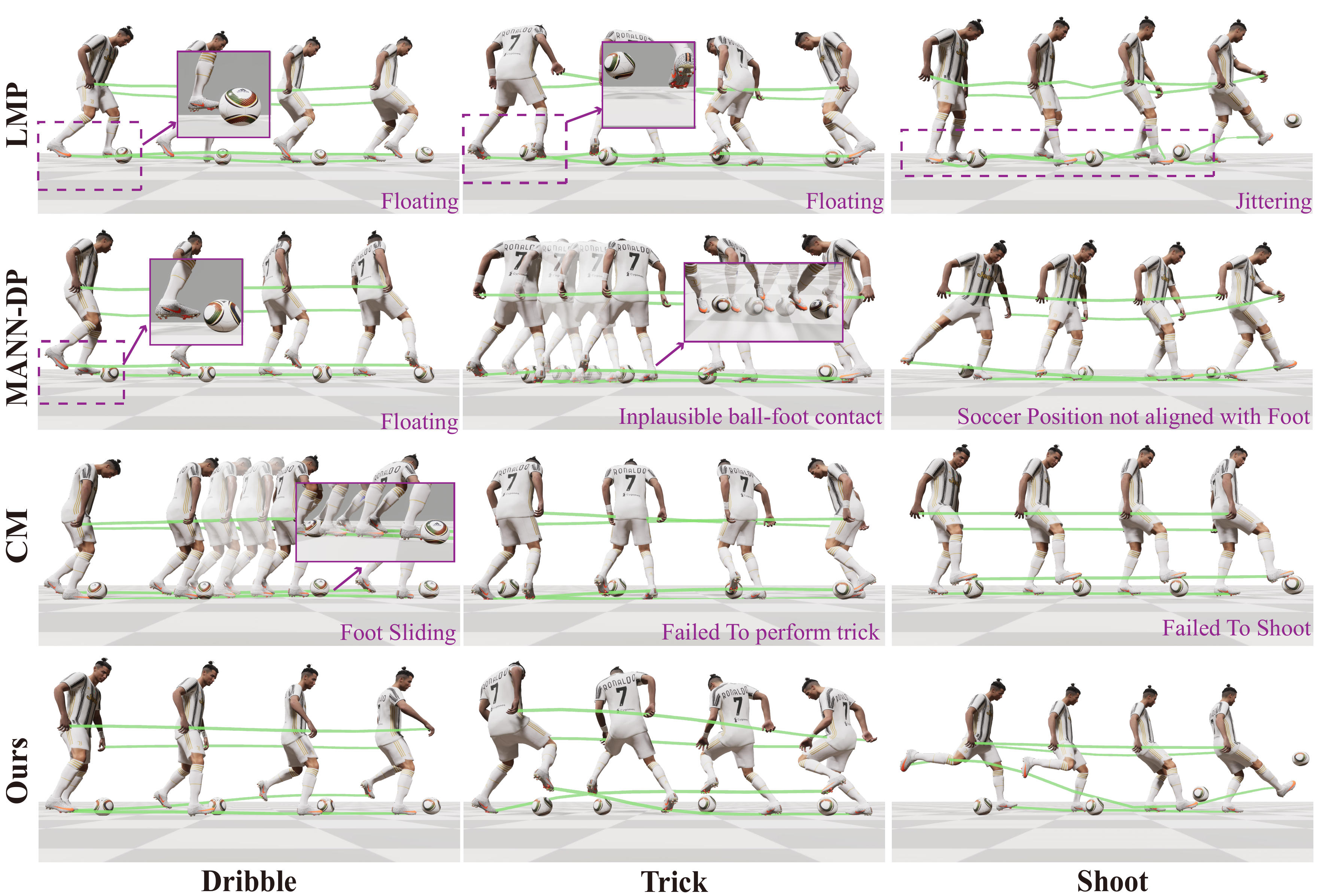}
    \caption{Qualitative comparison between our method and baseline methods including LMP~\cite{Starke2020}, MANN-DP~\cite{mann,deepphase} and CM~\cite{codebookmatching}. The green line represents the trajectories of the hands and feet. The motions generated by the baseline methods exhibit deficiencies in motion quality (such as foot sliding and skill accuracy). Our method significantly surpasses the baseline methods in terms of motion details. More qualitative results can be found in the supplementary video.}
    \label{fig:comparison}
    \
\end{figure*}

We set up a high-precision multi-view motion capture system comprising 16 OptiTrack Prime x13 cameras~\cite{Optitrack} in a 6m $\times$  7.5m $\times$ 2.5m capture area, recording at a rate of 240 frames per second. Our dataset includes recordings from 30 skilled soccer players, each demonstrating their unique playing styles. The final dataset contains over 10 hours of motion data, totaling 1.08 million frames. The human motion data adopts the SMPL model format~\cite{loper2015smpl}, while the soccer ball data includes both translational and rotational information. The dataset is subsequently downsampled to 30 fps for processing. All motion data is systematically organized into six categories: Dribble, Stand, Off-the-ball Move, Trick, Shoot, and Celebrate. 
The statistical details of the data are presented in Fig.~\ref{fig:dataset}. The Dribble category includes variations in speed, foot (left or right), and motion direction (straight-line and turning). The Trick motion features 5 specific soccer dribbling maneuvers. For the Shoot motion, due to spatial limitations in the indoor environment, the full ball trajectory was later reconstructed in Unity using physical simulations based on initially captured movements. Further details about the dataset are provided in the supplementary material.
\begin{table*}[t!]
\centering
\renewcommand\arraystretch{1.1}
\resizebox{0.8\linewidth}{!}{
\begin{tabular}
{ll|cccc|ccc}
\toprule[2pt]
\multicolumn{2}{l|}{} & \multicolumn{4}{c|}{Motion Quality} & \multicolumn{3}{c}{Condition Alignment}  \\
\cline{3-9}
 \multicolumn{2}{l|}{Method} & FID$\downarrow$ & Ft. Slid.$\downarrow$ & Accel.$\downarrow$ & Div.$\uparrow$  & Traj. Err.$\downarrow$ & Orient. Err.$\downarrow$ & Skill Acc.$\uparrow$ \\
\cline{1-9}
\multicolumn{2}{l|}{LMP~\cite{Starke2020}}        
& 0.3541 & 1.0678  & 1.6070 & 0.3980  & 4.1156 & 6.4932 & 73.3\% \\ 
\multicolumn{2}{l|}{MANN-DP~\cite{mann,deepphase}}    
& 0.3593 & 1.3507  & 1.5652 & 0.4753  & 4.0690 & 5.2985 & 69.1\% \\ 
\multicolumn{2}{l|}{CM~\cite{codebookmatching}}
& 0.2494 & 1.6498  & \textbf{1.1752} & 0.3520  & 3.1034 & 5.0663 & 52.9\% \\
\multicolumn{2}{l|}{Ours}
& \textbf{0.1813} & \textbf{0.8543}  & 1.1999 & \textbf{0.6177}  & \textbf{2.4132} & \textbf{4.9393} & \textbf{93.3\%} \\
\bottomrule[2pt]
\end{tabular}}
\caption{Quantitative comparison between our method and baseline methods including LMP~\cite{Starke2020}, MANN-DP~\cite{mann,deepphase} and CM~\cite{codebookmatching}. Our method outperforms the baseline methods in terms of both motion quality and condition alignment.}
\label{tab:comparison results}
\vspace{-1em} 
\end{table*}

\section{Experiments}
\label{sec:exp}
In this section, we validate the capability of our method through various experiments.

\myparagraph{Evaluation Dataset.}
Due to the lack of publicly available large-scale soccer datasets, we use our \textit{Soccer-X} dataset for both training and evaluation. The dataset is split into training and test sets with a 9:1 ratio. 

\begin{figure}[t]
    \centering
    \includegraphics[width=\linewidth]{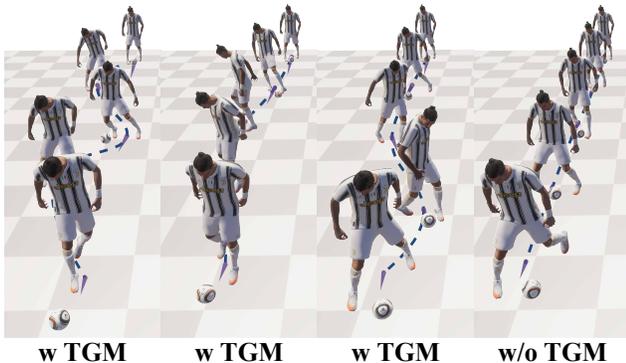}
    \caption{Qualitative evaluation of the Trajectory Generation Model (TGM). Given identical conditions, TGM enhances motion diversity. The dashed line represents the generated trajectory.}
    \label{fig:ablation trajectory}
\end{figure}

\myparagraph{Evaluation Metrics.}
To thoroughly assess the quality of generated motion, we adopt the following metrics, following previous work~\cite{camdm,tevet2023humanmotion}: (1) \textit{Fréchet Inception Distance} (FID), which measures the distance between the distribution of generated motions and that of the training data, providing insights into fidelity and similarity. (2) \textit{Foot Sliding Distance} (Ft. Slid.), quantifying the realism of the generated motion by measuring the distance (in meters) that the character’s toes move when the joint height falls below a specified threshold. (3) \textit{Acceleration} (Accel.), calculated as the mean per-joint acceleration (in centimeters per second), aids in assessing the smoothness of the generated motion. (4) \textit{Diversity} (Div.) measures the variance of each joint’s spatial locations over time for motions generated with the same control signals. The variance across all joints is averaged.

To further assess the alignment of generated motion with conditions, we employ (5) \textit{Trajectory Error} (Traj. Err.), which captures the angular difference (in degrees) between the expected and actual root movement directions; (6) \textit{Orientation Error} (Orient. Err.), the difference (in degrees) between expected and actual root orientations; and (7) \textit{Skill Accuracy} (Skill Acc.) quantifies the percentage of generated motions that align with the input style label. Following the approach~\cite{camdm}, we pre-trained a classifier to assess style accuracy.

\subsection{Comparison}
\myparagraph{Baseline Methods.}
We compare our method with several real-time character controller approaches. We benchmark against Local Motion Phase (LMP)~\cite{Starke2020}, a variant that combines Mixture of Experts from MANN~\cite{mann} with DeepPhase features~\cite{deepphase} (MANN-DP), and Categorical Codebook Matching (CM)~\cite{codebookmatching}. For a fair comparison, we use the same motion representation as our approach.

In comparison, we only focus on the quality of the soccer motions generated by different methods. Therefore, instead of applying trajectory generation, we use the trajectory inputs from the test set. To ensure a fair comparison, each method receives the same conditional input.

\begin{figure}[t]
    \centering
    \includegraphics[width=\linewidth]{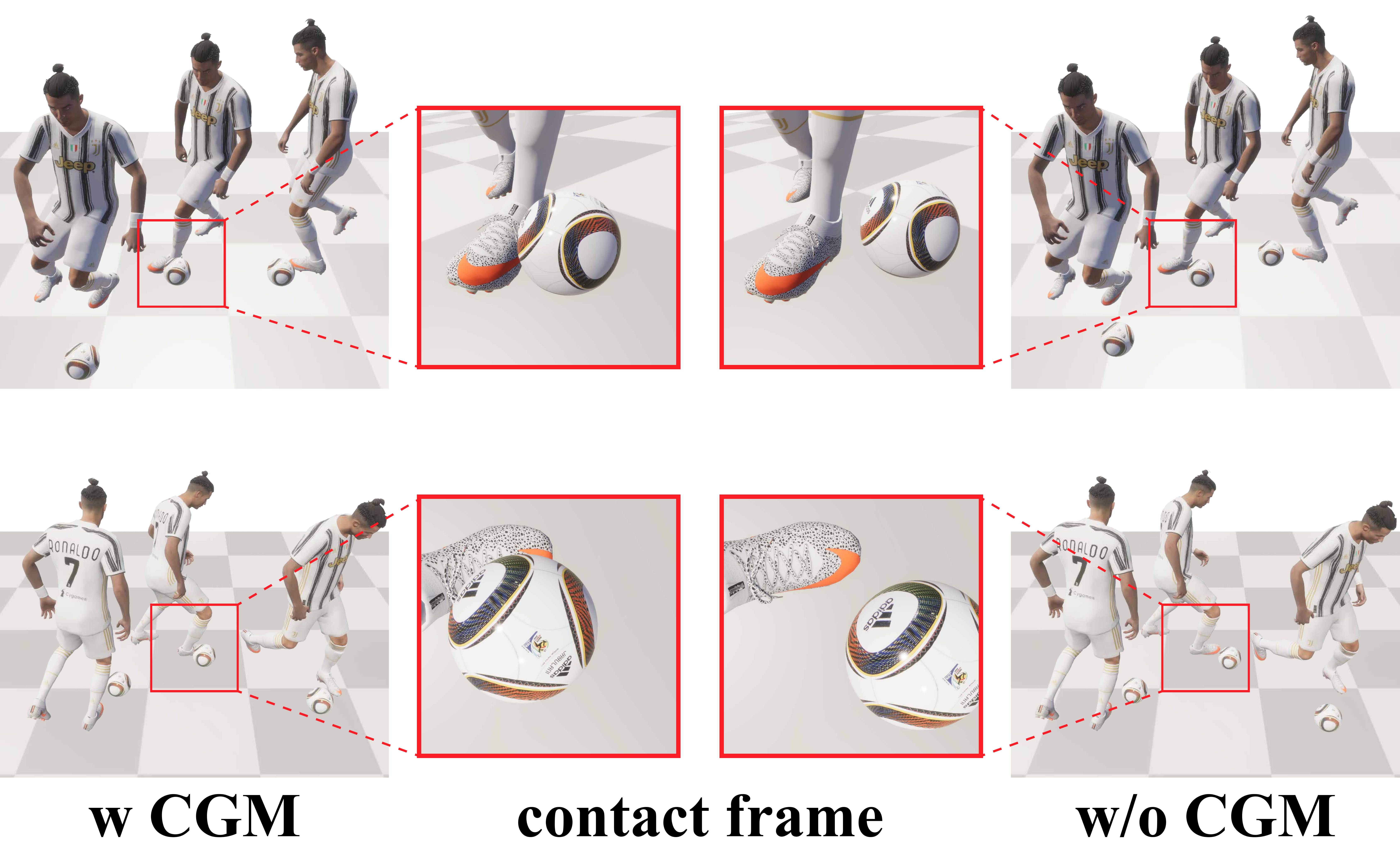}
    \caption{Qualitative evaluation of the Contact Guidance Module (CGM). Given identical conditions, CGM effectively prevents instances of missed contact when the ball changes direction. Contact frames represent points where the ball’s trajectory shifts.}
    \label{fig:contact guidance ablation}
\end{figure}

\myparagraph{Results.}
Tab.~\ref{tab:comparison results} presents the quantitative results of our experiments. Our method demonstrates superior motion quality and condition alignment. In terms of motion quality, our approach achieves the highest Diversity score and the lowest FID score, indicating that the generated motions are more diverse and better aligned with the distribution of soccer motion data. Additionally, our method exhibits the lowest foot sliding and comparable acceleration, further validating the realism of our generated motions. The CM~\cite{codebookmatching} method generates slightly smoother motions but suffers from significant foot sliding and inaccurate execution of soccer skills. Regarding condition alignment, our method consistently achieves the best performance across all metrics, showing the highest consistency between the generated motions and user inputs. Fig.~\ref{fig:comparison} provides the corresponding qualitative results of our experiments. Our method is capable of generating realistic and vivid football actions, while other methods often suffer from artifacts such as failure to generate the specified skill, foot sliding, jitter, or penetration issues.

\subsection{Ablation Study}

\myparagraph{Trajectory Generation Model (TGM).}
\label{sec:ablation TGM}
To analyze the effect of our trajectory generation model, we replace the generated trajectories with straight-line trajectories mapped from user control, denoted as \textit{w/o TGM}. In the quantitative results shown in Tab.~\ref{tab:ablation}, our full method achieves a significantly lower FID score and a higher Diversity score, indicating that incorporating the trajectory generation model produces more varied and realistic soccer motions. Qualitative results in Fig.~\ref{fig:ablation trajectory} reveal that, under the same user inputs, the motions generated with trajectory generation exhibit noticeably more complex trajectories and greater diversity.

\begin{table}[t]
\centering
\renewcommand\arraystretch{1.1}
\resizebox{1\linewidth}{!}{
\begin{tabular}{llcccc}
\toprule[2pt]
\multicolumn{2}{l}{} & FID$\downarrow$ & Ft.Slid.$\downarrow$ & Accel.$\downarrow$ & Div.$\uparrow$ \\
\cline{1-6}
\multicolumn{2}{l}{Ours w/o TGM}  
&  0.3646  &  \textbf{1.0031}  &  1.1973 & 2.4331    \\ 
\multicolumn{2}{l}{Ours w/o CGM}  
&  0.3704  &  1.0046  &  \textbf{1.1957} & 2.6908    \\ 
\multicolumn{2}{l}{Ours}   
&  \textbf{0.3580}  &  1.0049  &  1.2012 & \textbf{2.6925}    \\ 
\bottomrule[2pt]
\end{tabular}
}
\caption{Quantitative evaluation of Trajectory Generation Model (TGM) and Contact Guidance Module (CGM).}
\label{tab:ablation}
\vspace{-1em} 
\end{table}

\myparagraph{Contact Guidance Module (CGM).}
\label{sec:ablation CGM}
To verify the effectiveness of our contact guidance module, we compare our full model with a variant that excludes contact guidance during the diffusion process, denoted as \textit{w/o CGM}. As shown in Tab.~\ref{tab:ablation}, although adding contact guidance slightly reduces performance in terms of foot sliding and acceleration, the significant improvement in the FID metric demonstrates that contact guidance allows our method to generate motions that more closely resemble real-world human-ball interactions. Qualitative results in Fig.~\ref{fig:contact guidance ablation} demonstrate the enhancement of human-ball contact in the generated actions due to the addition of contact guidance.

\subsection{Runtime Analysis}
\label{sec:live demo}
In this section, we analyze the effects of different denoise steps and contact guidance employment strategies. We consider their impact on runtime efficiency and generation quality.

\myparagraph{Denoise Step.}
Consistent with previous findings~\cite{shi2023controllable, camdm}, we observe that directly training a diffusion model with fewer steps, rather than training a 1000-step model and applying the DDIM~\cite{song2022denoisingdiffusionimplicitmodels} sampling strategy, leads to better soccer motion generation. As shown in Table 3, using 8 denoise steps during training and testing achieves an optimal balance between inference speed and generation quality. Consequently, we apply 8 denoise steps in our live demo.

\myparagraph{Contact Guidance Strategy.}
Based on the 8 denoise steps setup, adding contact guidance to a maximum of 2 diffusion steps still enables real-time inference. To determine the optimal insertion point for contact guidance within the diffusion process, we compared five insertion strategies. In Tab.~\ref{tab:contact guidance scheduler}, \textit{Start $M$} denotes applying contact guidance in the initial $M$ denoise steps, while \textit{End $N$} applies contact guidance in the final $N$ steps of the diffusion process. As shown in Tab.~\ref{tab:contact guidance scheduler}, applying contact guidance in the last two steps (denoted as End 2) yields the lowest FID score, demonstrating superior generation quality. Therefore, in our live demo, we enable contact guidance only during the final two denoise steps.

\begin{table}[t!]
\centering
\renewcommand\arraystretch{1.1}
\resizebox{1\linewidth}{!}{
\begin{tabular}{lccccc}
\toprule[2pt]
\multicolumn{1}{l}{Steps} & 2 & 4 & 8 & 16 & 32 \\
\cline{1-6}
\multicolumn{1}{l}{Inf. Time}  
&  3ms  &  6ms  &  12ms  &  25ms  &  52ms   \\ 
\multicolumn{1}{l}{FID$\downarrow$}   
&  0.3954  &  0.3896  &  0.3704  &  0.3663  &  \textbf{0.3379}   \\ 
\multicolumn{1}{l}{Ft. Slid.$\downarrow$}   
&  1.2438  &  1.1471  &  \textbf{1.0046}  &  1.0434  &  1.0708   \\ 
\multicolumn{1}{l}{Accel.$\downarrow$}   
&  1.1882  &  \textbf{1.1751}  &  1.1957  &  1.1953  &  1.2007   \\ 
\bottomrule[2pt]
\end{tabular}
}
\caption{Evaluation of different denoise steps.}
\label{tab:denoise step}
\vspace{-1em} 
\end{table}

\begin{table}[t!]
\centering
\renewcommand\arraystretch{1.1}
\resizebox{1\linewidth}{!}{
\begin{tabular}{lcccccc}
\toprule[2pt]
\multicolumn{1}{l}{Schedule} & \multicolumn{1}{c}{Start 1} & \multicolumn{1}{c}{End 1} & \multicolumn{1}{c}{Start 2} & \multicolumn{1}{c}{End 2} & \multicolumn{1}{c}{Start 1, End 1} \\
\cline{1-6}
FID$\downarrow$ 
& 0.3642 & 0.3629 & 0.3672 & \textbf{0.3580} & 0.3661 \\ 
Ft. Slid.$\downarrow$ 
& \textbf{1.0035} & 1.0049 & 1.0046 & 1.0049 & 1.0041 \\ 
Accel.$\downarrow$ 
& 1.1939 & 1.2029 & \textbf{1.1931} & 1.2012 & 1.1984 \\ 
\bottomrule[2pt]
\end{tabular}
}
\caption{Evaluation of contact guidance employment strategy.}
\label{tab:contact guidance scheduler}
\vspace{-1em} 
\end{table}

\subsection{Limitation}
Although our SMGDiff can generate diverse and high-quality soccer motions, it still has several limitations. Firstly, although contact guidance is incorporated, the model lacks optimization for real-world physics during training and inference. Future work could benefit from refining the generated results within physics engines such as Isaac Gym~\cite{makoviychuk2021isaacgymhighperformance}. Second, we only consider interactions between the ball and the foot, neglecting other possible interactions between the ball and other parts of the body in real-life soccer scenarios. Third, the generated soccer motions do not involve interaction scenarios between different players. Therefore, a potential future direction is to integrate the diffusion model with a physics-based character animation method to generate physically realistic multi-person soccer motions.

\section{Conclusion}
In this paper, we present SMGDiff, a novel approach for generating diverse soccer motions including dribbling, playing tricks, shooting, etc. Our approach applies powerful generative networks to character animation, which enables real-time, user-controllable animation of vivid and varied soccer motions.
Our trajectory generation model utilizes a single-step diffusion model to transform high-level, coarse user inputs into refined trajectories, which indirectly enhances the diversity of the final generated motions.
Our soccer motion generation model is able to produce high-quality and condition-aligned motion sequences. 
Furthermore, the contact guidance module employs a specialized loss function to refine ball-foot interactions, resulting in more accurate and realistic soccer animations. 
The extensive experimental results demonstrate the effectiveness of SMGDiff in real-time, controllable soccer motion generation tasks. We believe that our approach and dataset represent a significant step toward the field of character animation, with potential applications in gaming and VR/AR.
{
    \small
    \bibliographystyle{ieeenat_fullname}
    \bibliography{main}

\begin{thebibliography}{81}
\providecommand{\natexlab}[1]{#1}
\providecommand{\url}[1]{\texttt{#1}}
\expandafter\ifx\csname urlstyle\endcsname\relax
  \providecommand{\doi}[1]{doi: #1}\else
  \providecommand{\doi}{doi: \begingroup \urlstyle{rm}\Url}\fi

\bibitem[Alexanderson et~al.(2022)Alexanderson, Nagy, Beskow, and Henter]{Alexanderson2022ListenDA}
Simon Alexanderson, Rajmund Nagy, Jonas Beskow, and Gustav~Eje Henter.
\newblock Listen, denoise, action! audio-driven motion synthesis with diffusion models.
\newblock \emph{ACM Trans. Graph.}, 2022.

\bibitem[Ao et~al.(2023)Ao, Zhang, and Liu]{ao2023gesturediffuclip}
Tenglong Ao, Zeyi Zhang, and Libin Liu.
\newblock Gesturediffuclip: Gesture diffusion model with clip latents.
\newblock \emph{ACM Transactions on Graphics (TOG)}, 42\penalty0 (4):\penalty0 1--18, 2023.

\bibitem[Arikan and Forsyth(2002)]{ArikanForsyth2002}
Okan Arikan and David~A. Forsyth.
\newblock Interactive motion generation from examples.
\newblock \emph{ACM Transactions on Graphics (TOG)}, 2002.

\bibitem[Chen et~al.(2023{\natexlab{a}})Chen, Wei, Lu, Zhu, Yao, Xiao, and Chen]{Chen2023DiffusionTalkerPA}
Peng Chen, Xiaobao Wei, Ming Lu, Yitong Zhu, Naiming Yao, Xingyu Xiao, and Hui Chen.
\newblock Diffusiontalker: Personalization and acceleration for speech-driven 3d face diffuser.
\newblock \emph{arXiv preprint arXiv:2311.16565}, 2023{\natexlab{a}}.

\bibitem[Chen et~al.(2024)Chen, Shi, Huang, Tan, Komura, and Chen]{camdm}
Rui Chen, Mingyi Shi, Shaoli Huang, Ping Tan, Taku Komura, and Xuelin Chen.
\newblock Taming diffusion probabilistic models for character control.
\newblock In \emph{SIGGRAPH}, 2024.

\bibitem[Chen et~al.(2023{\natexlab{b}})Chen, Jiang, Liu, Huang, Fu, Chen, and Yu]{chen2023executing}
Xin Chen, Biao Jiang, Wen Liu, Zilong Huang, Bin Fu, Tao Chen, and Gang Yu.
\newblock Executing your commands via motion diffusion in latent space.
\newblock In \emph{Proceedings of the IEEE/CVF Conference on Computer Vision and Pattern Recognition}, pages 18000--18010, 2023{\natexlab{b}}.

\bibitem[Cho et~al.(2021)Cho, Kim, Park, Park, and Noh]{motionrecom}
Kyungmin Cho, Chaelin Kim, Jungjin Park, Joonkyu Park, and Junyong Noh.
\newblock Motion recommendation for online character control.
\newblock \emph{ACM Transactions on Graphics (TOG)}, 2021.

\bibitem[Clavet(2016)]{clavet2016motion}
Simon Clavet.
\newblock Motion matching and the road to next-gen animation.
\newblock In \emph{Proceedings of the Game Developers Conference (GDC)}, 2016.

\bibitem[Diller and Dai(2024)]{CG-HOI}
Christian Diller and Angela Dai.
\newblock Cg-hoi: Contact-guided 3d human-object interaction generation.
\newblock In \emph{2024 IEEE/CVF Conference on Computer Vision and Pattern Recognition (CVPR)}, 2024.

\bibitem[{Electronic Arts}(2024)]{eaSportsFc}
{Electronic Arts}.
\newblock {EA SPORTS FC - Official Website}.
\newblock {\url{https://www.ea.com/games/ea-sports-fc}}, 2024.

\bibitem[Fragkiadaki et~al.(2015)Fragkiadaki, Levine, Felsen, and Malik]{fragkiadaki2015recurrent}
Katerina Fragkiadaki, Sergey Levine, Panna Felsen, and Jitendra Malik.
\newblock Recurrent network models for human dynamics.
\newblock In \emph{Proceedings of the IEEE International Conference on Computer Vision (ICCV)}, pages 4346--4354. IEEE, 2015.

\bibitem[Ghosh et~al.(2023)Ghosh, Dabral, Golyanik, Theobalt, and Slusallek]{ghosh2023imos}
Anindita Ghosh, Rishabh Dabral, Vladislav Golyanik, Christian Theobalt, and Philipp Slusallek.
\newblock Imos: Intent-driven full-body motion synthesis for human-object interactions.
\newblock \emph{Comput. Graph. Forum}, 42\penalty0 (2):\penalty0 1--12, 2023.

\bibitem[Guo et~al.(2022)Guo, Zou, Zuo, Wang, Ji, Li, and Cheng]{guo2022generating}
Chuan Guo, Shihao Zou, Xinxin Zuo, Sen Wang, Wei Ji, Xingyu Li, and Li Cheng.
\newblock Generating diverse and natural 3d human motions from text.
\newblock In \emph{Proceedings of the IEEE/CVF Conference on Computer Vision and Pattern Recognition}, pages 5152--5161, 2022.

\bibitem[Hassan et~al.(2023)Hassan, Guo, Wang, Black, Fidler, and Peng]{hassan2023synthesizingphysicalcharactersceneinteractions}
Mohamed Hassan, Yunrong Guo, Tingwu Wang, Michael Black, Sanja Fidler, and Xue~Bin Peng.
\newblock Synthesizing physical character-scene interactions, 2023.

\bibitem[Ho et~al.(2020{\natexlab{a}})Ho, Jain, and Abbeel]{DDPM}
Jonathan Ho, Ajay Jain, and Pieter Abbeel.
\newblock Denoising diffusion probabilistic models.
\newblock In \emph{Advances in Neural Information Processing Systems}, pages 6840--6851. Curran Associates, Inc., 2020{\natexlab{a}}.

\bibitem[Ho et~al.(2020{\natexlab{b}})Ho, Jain, and Abbeel]{findlay2022denoising}
Jonathan Ho, Ajay Jain, and Pieter Abbeel.
\newblock Denoising diffusion probabilistic models.
\newblock In \emph{Advances in Neural Information Processing Systems}, 2020{\natexlab{b}}.

\bibitem[Ho et~al.(2020{\natexlab{c}})Ho, Jain, and Abbeel]{ho2020denoising}
Jonathan Ho, Ajay Jain, and Pieter Abbeel.
\newblock Denoising diffusion probabilistic models.
\newblock In \emph{Advances in Neural Information Processing Systems}, pages 6840--6851, 2020{\natexlab{c}}.

\bibitem[Holden et~al.(2015)Holden, Saito, Komura, and Joyce]{holden2015learning}
Daniel Holden, Jun Saito, Taku Komura, and Thomas Joyce.
\newblock Learning motion manifolds with convolutional autoencoders.
\newblock In \emph{SIGGRAPH Asia 2015 Technical Briefs}, page~18. ACM, 2015.

\bibitem[Holden et~al.(2016)Holden, Saito, and Komura]{holden2016deep}
Daniel Holden, Jun Saito, and Taku Komura.
\newblock A deep learning framework for character motion synthesis and editing.
\newblock \emph{ACM Transactions on Graphics (ToG)}, 2016.

\bibitem[Holden et~al.(2017)Holden, Komura, and Saito]{pfnn}
Daniel Holden, Taku Komura, and Jun Saito.
\newblock Phase-functioned neural networks for character control.
\newblock \emph{ACM Trans. Graph.}, 2017.

\bibitem[Holden et~al.(2020{\natexlab{a}})Holden, Kanoun, Perepichka, and Popa]{holden2020learned}
Daniel Holden, Oussama Kanoun, Maksym Perepichka, and Tiberiu Popa.
\newblock Learned motion matching.
\newblock \emph{ACM Transactions on Graphics (ToG)}, 2020{\natexlab{a}}.

\bibitem[Holden et~al.(2020{\natexlab{b}})Holden, Kanoun, Perepichka, and Popa]{learnedmm}
Daniel Holden, Oussama Kanoun, Maksym Perepichka, and Tiberiu Popa.
\newblock Learned motion matching.
\newblock \emph{ACM Trans. Graph.}, 2020{\natexlab{b}}.

\bibitem[Hong et~al.(2019)Hong, Han, Cho, Shin, and Noh]{10.1145/3306346.3322963}
Seokpyo Hong, Daseong Han, Kyungmin Cho, Joseph~S. Shin, and Junyong Noh.
\newblock Physics-based full-body soccer motion control for dribbling and shooting.
\newblock \emph{ACM Trans. Graph.}, 38\penalty0 (4), 2019.

\bibitem[Huang et~al.(2023)Huang, Wang, Li, Jia, Liu, Zhu, Liang, and Zhu]{Huang_2023_diffu_based_gen}
Siyuan Huang, Zan Wang, Puhao Li, Baoxiong Jia, Tengyu Liu, Yixin Zhu, Wei Liang, and Song-Chun Zhu.
\newblock Diffusion-based generation, optimization, and planning in 3d scenes.
\newblock In \emph{Proceedings of the IEEE/CVF Conference on Computer Vision and Pattern Recognition (CVPR)}, pages 16750--16761, 2023.

\bibitem[Karunratanakul et~al.(2023)Karunratanakul, Preechakul, Suwajanakorn, and Tang]{gmd}
Korrawe Karunratanakul, Konpat Preechakul, Supasorn Suwajanakorn, and Siyu Tang.
\newblock Guided motion diffusion for controllable human motion synthesis.
\newblock In \emph{Proceedings of the IEEE/CVF International Conference on Computer Vision}, pages 2151--2162, 2023.

\bibitem[Karunratanakul et~al.(2024)Karunratanakul, Preechakul, Aksan, Beeler, Suwajanakorn, and Tang]{odn}
Korrawe Karunratanakul, Konpat Preechakul, Emre Aksan, Thabo Beeler, Supasorn Suwajanakorn, and Siyu Tang.
\newblock Optimizing diffusion noise can serve as universal motion priors.
\newblock In \emph{Proceedings of the IEEE/CVF Conference on Computer Vision and Pattern Recognition (CVPR)}, 2024.

\bibitem[Kim et~al.(2023)Kim, Kim, and Choi]{kim2023flame}
Jihoon Kim, Jiseob Kim, and Sungjoon Choi.
\newblock Flame: Free-form language-based motion synthesis \& editing.
\newblock In \emph{Proceedings of the AAAI Conference on Artificial Intelligence}, pages 8255--8263, 2023.

\bibitem[Kovar et~al.(2002)Kovar, Gleicher, and Pighin]{Kovar2002}
Lucas Kovar, Michael Gleicher, and Frédéric Pighin.
\newblock Motion graphs.
\newblock \emph{ACM Transactions on Graphics}, 21\penalty0 (3):\penalty0 473--482, 2002.

\bibitem[Lau and Kuffner(2005)]{lau2005behavior}
Manfred Lau and James~J. Kuffner.
\newblock Behavior planning for character animation.
\newblock In \emph{Proceedings of the Symposium on Computer Animation (SCA)}, 2005.

\bibitem[Lee et~al.(2018)Lee, Lee, and Lee]{lee2018interactive}
Kyungho Lee, Seyoung Lee, and Jehee Lee.
\newblock Interactive character animation by learning multi-objective control.
\newblock \emph{ACM Transactions on Graphics (TOG)}, 2018.

\bibitem[Lee et~al.(2021)Lee, Min, Lee, and Lee]{lee2021learning}
Kyungho Lee, Sehee Min, Sunmin Lee, and Jehee Lee.
\newblock Learning time-critical responses for interactive character control.
\newblock \emph{ACM Transactions on Graphics (ToG)}, 40\penalty0 (4):\penalty0 Article 147, 2021.

\bibitem[Li et~al.(2023{\natexlab{a}})]{Li2023ObjectMG}
Jiaman Li et~al.
\newblock Object motion guided human motion synthesis.
\newblock \emph{ACM Trans. Graph.}, 2023{\natexlab{a}}.

\bibitem[Li et~al.(2023{\natexlab{b}})Li, Wang, Loy, and Dai]{li2023task}
Quanzhou Li, Jingbo Wang, Chen~Change Loy, and Bo Dai.
\newblock Task-oriented human-object interactions generation with implicit neural representations.
\newblock \emph{CoRR}, abs/2303.13129, 2023{\natexlab{b}}.

\bibitem[Li et~al.(2017)Li, Zhou, Xiao, He, Huang, and Li]{li2017auto}
Zimo Li, Yi Zhou, Shuangjiu Xiao, Chong He, Zeng Huang, and Hao Li.
\newblock Auto-conditioned recurrent networks for extended complex human motion synthesis.
\newblock \emph{arXiv preprint arXiv:1707.05363}, 2017.

\bibitem[Liang et~al.(2023)Liang, Zhang, Li, Yu, and Xu]{liang2023intergen}
Han Liang, Wenqian Zhang, Wenxuan Li, Jingyi Yu, and Lan Xu.
\newblock Intergen: Diffusion-based multi-human motion generation under complex interactions.
\newblock \emph{International Journal of Computer Vision}, 2023.

\bibitem[Liang et~al.(2024)Liang, Bao, Zhang, Ren, Xu, Yang, Chen, Yu, and Xu]{Liang_2024_CVPR}
Han Liang, Jiacheng Bao, Ruichi Zhang, Sihan Ren, Yuecheng Xu, Sibei Yang, Xin Chen, Jingyi Yu, and Lan Xu.
\newblock Omg: Towards open-vocabulary motion generation via mixture of controllers.
\newblock In \emph{Proceedings of the IEEE/CVF Conference on Computer Vision and Pattern Recognition (CVPR)}, pages 482--493, 2024.

\bibitem[Libin~Liu(August 2018)]{Hodgins:2017:DOE}
Jessica~Hodgins Libin~Liu.
\newblock Learning basketball dribbling skills using trajectory optimization and deep reinforcement learning.
\newblock \emph{ACM Transactions on Graphics}, August 2018.

\bibitem[Liu et~al.(2021)Liu, Lever, Wang, Merel, Eslami, Hennes, Czarnecki, Tassa, Omidshafiei, Abdolmaleki, Siegel, Hasenclever, Marris, Tunyasuvunakool, Song, Wulfmeier, Muller, Haarnoja, Tracey, Tuyls, Graepel, and Heess]{liu2021motorcontrolteamplay}
Siqi Liu, Guy Lever, Zhe Wang, Josh Merel, S.~M.~Ali Eslami, Daniel Hennes, Wojciech~M. Czarnecki, Yuval Tassa, Shayegan Omidshafiei, Abbas Abdolmaleki, Noah~Y. Siegel, Leonard Hasenclever, Luke Marris, Saran Tunyasuvunakool, H.~Francis Song, Markus Wulfmeier, Paul Muller, Tuomas Haarnoja, Brendan~D. Tracey, Karl Tuyls, Thore Graepel, and Nicolas Heess.
\newblock From motor control to team play in simulated humanoid football, 2021.

\bibitem[Loper et~al.(2015{\natexlab{a}})Loper, Mahmood, Romero, Pons-Moll, and Black]{SMPL}
Matthew Loper, Naureen Mahmood, Javier Romero, Gerard Pons-Moll, and Michael~J. Black.
\newblock Smpl: a skinned multi-person linear model.
\newblock \emph{ACM Trans. Graph.}, 34\penalty0 (6), 2015{\natexlab{a}}.

\bibitem[Loper et~al.(2015{\natexlab{b}})Loper, Mahmood, Romero, Pons-Moll, and Black]{loper2015smpl}
Matthew Loper, Naureen Mahmood, Javier Romero, Gerard Pons-Moll, and Michael~J Black.
\newblock Smpl: A skinned multi-person linear model.
\newblock \emph{ACM Transactions on Graphics (TOG)}, 34\penalty0 (6):\penalty0 1--16, 2015{\natexlab{b}}.

\bibitem[Mach and Zhuravlov(2021)]{mach2021motion}
Michal Mach and Maksym Zhuravlov.
\newblock Motion matching in 'the last of us part ii', 2021.
\newblock \url{https://www.gdcvault.com/play/1027118/Motion-Matching-in-The-Last}.

\bibitem[Makoviychuk et~al.(2021)Makoviychuk, Wawrzyniak, Guo, Lu, Storey, Macklin, Hoeller, Rudin, Allshire, Handa, and State]{makoviychuk2021isaacgymhighperformance}
Viktor Makoviychuk, Lukasz Wawrzyniak, Yunrong Guo, Michelle Lu, Kier Storey, Miles Macklin, David Hoeller, Nikita Rudin, Arthur Allshire, Ankur Handa, and Gavriel State.
\newblock Isaac gym: High performance gpu-based physics simulation for robot learning, 2021.

\bibitem[Mason et~al.(2022)Mason, Starke, and Komura]{mason2022real}
Ian Mason, Sebastian Starke, and Taku Komura.
\newblock Real-time style modelling of human locomotion via feature-wise transformations and local motion phases.
\newblock \emph{Proc. ACM Comput. Graph. Interact. Tech.}, 2022.

\bibitem[Merel et~al.(2020)Merel, Tunyasuvunakool, Ahuja, Tassa, Hasenclever, Pham, Erez, Wayne, and Heess]{merel2020catchcarryreusable}
Josh Merel, Saran Tunyasuvunakool, Arun Ahuja, Yuval Tassa, Leonard Hasenclever, Vu Pham, Tom Erez, Greg Wayne, and Nicolas Heess.
\newblock Catch \& carry: Reusable neural controllers for vision-guided whole-body tasks, 2020.

\bibitem[Nichol et~al.(2022)Nichol, Dhariwal, Ramesh, Shyam, Mishkin, McGrew, Sutskever, and Chen]{nichol2022glide}
Alexander~Quinn Nichol, Prafulla Dhariwal, Aditya Ramesh, Pranav Shyam, Pamela Mishkin, Bob McGrew, Ilya Sutskever, and Mark Chen.
\newblock {GLIDE}: Towards photorealistic image generation and editing with text-guided diffusion models.
\newblock In \emph{International Conference on Machine Learning (ICML)}, 2022.

\bibitem[Optitrack()]{Optitrack}
Optitrack.
\newblock {NaturalPoint, Inc. Motion Capture Systems.}
\newblock \url{https://optitrack.com/}, 2022.

\bibitem[Peng et~al.(2019)Peng, Chang, Zhang, Abbeel, and Levine]{peng2019mcplearningcomposablehierarchical}
Xue~Bin Peng, Michael Chang, Grace Zhang, Pieter Abbeel, and Sergey Levine.
\newblock Mcp: Learning composable hierarchical control with multiplicative compositional policies, 2019.

\bibitem[Peng et~al.(2021{\natexlab{a}})Peng, Ma, Abbeel, Levine, and Kanazawa]{Peng_2021}
Xue~Bin Peng, Ze Ma, Pieter Abbeel, Sergey Levine, and Angjoo Kanazawa.
\newblock Amp: adversarial motion priors for stylized physics-based character control.
\newblock \emph{ACM Transactions on Graphics}, 2021{\natexlab{a}}.

\bibitem[Peng et~al.(2021{\natexlab{b}})Peng, Ma, Abbeel, Levine, and Kanazawa]{peng2021amp}
Xue~Bin Peng, Ze Ma, Pieter Abbeel, Sergey Levine, and Angjoo Kanazawa.
\newblock Amp: Adversarial motion priors for stylized physics-based character control.
\newblock \emph{ACM Transactions on Graphics (ToG)}, 40\penalty0 (4):\penalty0 1--20, 2021{\natexlab{b}}.

\bibitem[Ramesh et~al.(2022)Ramesh, Dhariwal, Nichol, Chu, and Chen]{Ramesh2022HierarchicalTI}
Aditya Ramesh, Prafulla Dhariwal, Alex Nichol, Casey Chu, and Mark Chen.
\newblock Hierarchical text-conditional image generation with clip latents.
\newblock \emph{ArXiv}, 2022.

\bibitem[Saharia et~al.(2022)Saharia, Chan, Saxena, Li, Whang, Denton, Ghasemipour, Lopes, Ayan, Salimans, and et~al.]{saharia2022photorealistic}
Chitwan Saharia, William Chan, Saurabh Saxena, Lala Li, Jay Whang, Emily~L Denton, Kamyar Ghasemipour, Raphael~Gontijo Lopes, Burcu~Karagol Ayan, Tim Salimans, and et al.
\newblock Photorealistic text-to-image diffusion models with deep language understanding.
\newblock In \emph{Advances in Neural Information Processing Systems (NeurIPS)}, 2022.

\bibitem[Shi et~al.(2023)Shi, Wang, Jiang, and Dai]{shi2023controllable}
Yi Shi, Jingbo Wang, Xuekun Jiang, and Bo Dai.
\newblock Controllable motion diffusion model.
\newblock \emph{CoRR}, 2023.

\bibitem[Song et~al.(2022)Song, Meng, and Ermon]{song2022denoisingdiffusionimplicitmodels}
Jiaming Song, Chenlin Meng, and Stefano Ermon.
\newblock Denoising diffusion implicit models, 2022.

\bibitem[Starke et~al.(2019{\natexlab{a}})Starke, Zhang, Komura, and Saito]{10.1145/3355089.3356505}
Sebastian Starke, He Zhang, Taku Komura, and Jun Saito.
\newblock Neural state machine for character-scene interactions.
\newblock \emph{ACM Trans. Graph.}, 38\penalty0 (6), 2019{\natexlab{a}}.

\bibitem[Starke et~al.(2019{\natexlab{b}})Starke, Zhang, Komura, and Saito]{starke2019neural}
Sebastian Starke, He Zhang, Taku Komura, and Jun Saito.
\newblock Neural state machine for character-scene interactions.
\newblock \emph{ACM Transactions on Graphics (ToG)}, 38\penalty0 (6):\penalty0 209:1--209:10, 2019{\natexlab{b}}.

\bibitem[Starke et~al.(2020{\natexlab{a}})Starke, Zhao, Komura, and Zaman]{10.1145/3386569.3392450}
Sebastian Starke, Yiwei Zhao, Taku Komura, and Kazi Zaman.
\newblock Local motion phases for learning multi-contact character movements.
\newblock \emph{ACM Trans. Graph.}, 39\penalty0 (4), 2020{\natexlab{a}}.

\bibitem[Starke et~al.(2020{\natexlab{b}})Starke, Zhao, Komura, and Zaman]{Starke2020}
Sebastian Starke, Yiwei Zhao, Taku Komura, and Kazi Zaman.
\newblock Local motion phases for learning multi-contact character movements.
\newblock \emph{ACM Transactions on Graphics (SIGGRAPH)}, 2020{\natexlab{b}}.

\bibitem[Starke et~al.(2021)Starke, Zhao, Zinno, and Komura]{starke2021neural}
Sebastian Starke, Yiwei Zhao, Fabio Zinno, and Taku Komura.
\newblock Neural animation layering for synthesizing martial arts movements.
\newblock \emph{ACM Transactions on Graphics (TOG)}, 2021.

\bibitem[Starke et~al.(2022)Starke, Mason, and Komura]{deepphase}
Sebastian Starke, Ian Mason, and Taku Komura.
\newblock Deepphase: Periodic autoencoders for learning motion phase manifolds.
\newblock \emph{ACM Transactions on Graphics (ToG)}, 2022.

\bibitem[Starke et~al.(2024)Starke, Starke, He, Komura, and Ye]{codebookmatching}
Sebastian Starke, Paul Starke, Nicky He, Taku Komura, and Yuting Ye.
\newblock Categorical codebook matching for embodied character controllers.
\newblock \emph{ACM Trans. Graph.}, 2024.

\bibitem[Tevet et~al.(2023)Tevet, Raab, Gordon, Shafir, Cohen-Or, and Bermano]{tevet2023humanmotion}
Guy Tevet, Sigal Raab, Brian Gordon, Yonatan Shafir, Daniel Cohen-Or, and Amit~H. Bermano.
\newblock Human motion diffusion model.
\newblock In \emph{International Conference on Learning Representations (ICLR)}, 2023.

\bibitem[Tevet et~al.(2024)Tevet, Raab, Cohan, Reda, Luo, Peng, Bermano, and van~de Panne]{tevet2024closd}
Guy Tevet, Sigal Raab, Setareh Cohan, Daniele Reda, Zhengyi Luo, Xue~Bin Peng, Amit~H Bermano, and Michiel van~de Panne.
\newblock Closd: Closing the loop between simulation and diffusion for multi-task character control.
\newblock \emph{arXiv preprint arXiv:2410.03441}, 2024.

\bibitem[Tseng et~al.(2023)Tseng, Castellon, and Liu]{tseng2022edge}
Jonathan Tseng, Rodrigo Castellon, and C.~Karen Liu.
\newblock Edge: Editable dance generation from music.
\newblock In \emph{2023 IEEE/CVF Conference on Computer Vision and Pattern Recognition (CVPR)}, 2023.

\bibitem[Xie et~al.(2023{\natexlab{a}})Xie, Jampani, Zhong, Sun, and Jiang]{xie2023omnicontrol}
Yiming Xie, Varun Jampani, Lei Zhong, Deqing Sun, and Huaizu Jiang.
\newblock Omnicontrol: Control any joint at any time for human motion generation.
\newblock In \emph{The Twelfth International Conference on Learning Representations}, 2023{\natexlab{a}}.

\bibitem[Xie et~al.(2022)Xie, Starke, Ling, and van~de Panne]{2022-Soccer-Juggle}
Zhaoming Xie, Sebastian Starke, Hung~Yu Ling, and Michiel van~de Panne.
\newblock Learning soccer juggling skills with layer-wise mixture-of-experts.
\newblock In \emph{ACM SIGGRAPH 2022 Conference Proceedings}, 2022.

\bibitem[Xie et~al.(2023{\natexlab{b}})Xie, Tseng, Starke, van~de Panne, and Liu]{xie2023hierarchicalplanningcontrolbox}
Zhaoming Xie, Jonathan Tseng, Sebastian Starke, Michiel van~de Panne, and C.~Karen Liu.
\newblock Hierarchical planning and control for box loco-manipulation, 2023{\natexlab{b}}.

\bibitem[Yang et~al.(2024)Yang, Ding, Cai, Yu, Wang, and Shi]{yang2023dsg}
Lingxiao Yang, Shutong Ding, Yifan Cai, Jingyi Yu, Jingya Wang, and Ye Shi.
\newblock Guidance with spherical gaussian constraint for conditional diffusion.
\newblock In \emph{International Conference on Machine Learning}, 2024.

\bibitem[Yao et~al.(2022)Yao, Song, Chen, and Liu]{ControlVAE}
Heyuan Yao, Zhenhua Song, Baoquan Chen, and Libin Liu.
\newblock Controlvae: Model-based learning of generative controllers for physics-based characters.
\newblock \emph{ACM Trans. Graph.}, 2022.

\bibitem[Yao et~al.(2023)Yao, Song, Zhou, Ao, Chen, and Liu]{yao2023moconvqunifiedphysicsbasedmotion}
Heyuan Yao, Zhenhua Song, Yuyang Zhou, Tenglong Ao, Baoquan Chen, and Libin Liu.
\newblock Moconvq: Unified physics-based motion control via scalable discrete representations.
\newblock In \emph{ACM SIGGRAPH 2024 Journal Proceedings}, 2023.

\bibitem[Ye et~al.(2022)Ye, Liu, Hu, and Xia]{ye2022neural3points}
Yongjing Ye, Libin Liu, Lei Hu, and Shihong Xia.
\newblock Neural3points: Learning to generate physically realistic full-body motion for virtual reality users.
\newblock In \emph{Computer Graphics Forum}. Wiley Online Library, 2022.

\bibitem[Yuan et~al.(2023)Yuan, Song, Iqbal, Vahdat, and Kautz]{yuan2023physdiff}
Ye Yuan, Jiaming Song, Umar Iqbal, Arash Vahdat, and Jan Kautz.
\newblock Physdiff: Physics-guided human motion diffusion model.
\newblock In \emph{Proceedings of the IEEE/CVF International Conference on Computer Vision}, pages 16010--16021, 2023.

\bibitem[Zhang et~al.(2018)Zhang, Starke, Komura, and Saito]{mann}
He Zhang, Sebastian Starke, Taku Komura, and Jun Saito.
\newblock Mode-adaptive neural networks for quadruped motion control.
\newblock \emph{ACM Trans. Graph.}, 2018.

\bibitem[Zhang et~al.(2023)Zhang, Rao, and Agrawala]{Zhang_2023_controlnet}
Lvmin Zhang, Anyi Rao, and Maneesh Agrawala.
\newblock Adding conditional control to text-to-image diffusion models.
\newblock In \emph{Proceedings of the IEEE/CVF International Conference on Computer Vision (ICCV)}, pages 3836--3847, 2023.

\bibitem[Zhang et~al.(2022)Zhang, Cai, Pan, Hong, Guo, Yang, and Liu]{zhang2022motiondiffuse}
Mingyuan Zhang, Zhongang Cai, Liang Pan, Fangzhou Hong, Xinying Guo, Lei Yang, and Ziwei Liu.
\newblock Motiondiffuse: Text-driven human motion generation with diffusion model.
\newblock \emph{IEEE Transactions on Pattern Analysis \& Machine Intelligence}, 2022.

\bibitem[Zhao et~al.(2024{\natexlab{a}})Zhao, Li, and Tang]{Zhao:DART:2024}
Kaifeng Zhao, Gen Li, and Siyu Tang.
\newblock Dart: A diffusion-based autoregressive motion model for real-time text-driven motion control, 2024{\natexlab{a}}.

\bibitem[Zhao et~al.(2023)Zhao, Liu, Ren, and Dai]{zhao_2023_Modiff}
M. Zhao, M. Liu, B. Ren, and S. Dai.
\newblock Modiff: Action-conditioned 3d motion generation with denoising diffusion probabilistic models.
\newblock \emph{arXiv preprint arXiv:2301.03949}, 2023.

\bibitem[Zhao et~al.(2024{\natexlab{b}})Zhao, Liu, Ren, and Dai]{Zhao2024DenoisingDP}
M. Zhao, M. Liu, B. Ren, and S. Dai.
\newblock Denoising diffusion probabilistic models for action-conditioned 3d motion generation.
\newblock In \emph{ICASSP 2024 - 2024 IEEE International Conference on Acoustics, Speech and Signal Processing (ICASSP)}, 2024{\natexlab{b}}.

\bibitem[Zhao et~al.(2024{\natexlab{c}})Zhao, Long, Zhang, Qin, Liang, Zhang, Zhang, Yu, and Xu]{zhao2024media2face}
Qingcheng Zhao, Pengyu Long, Qixuan Zhang, Dafei Qin, Han Liang, Longwen Zhang, Yingliang Zhang, Jingyi Yu, and Lan Xu.
\newblock Media2face: Co-speech facial animation generation with multi-modality guidance.
\newblock In \emph{ACM SIGGRAPH 2024 Conference Papers}, 2024{\natexlab{c}}.

\bibitem[Zhou et~al.(2024)Zhou, Dou, Cao, Liao, Wang, Wang, Liu, Komura, Wang, and Liu]{zhou2023emdm}
Wenyang Zhou, Zhiyang Dou, Zeyu Cao, Zhouyingcheng Liao, Jingbo Wang, Wenjia Wang, Yuan Liu, Taku Komura, Wenping Wang, and Lingjie Liu.
\newblock Emdm: Efficient motion diffusion model for fast, high-quality motion generation.
\newblock \emph{ECCV}, 2024.

\bibitem[Zhou et~al.(2018)Zhou, Barnes, Lu, Yang, and Li]{R6D}
Yi Zhou, Connelly Barnes, Jingwan Lu, Jimei Yang, and Hao Li.
\newblock On the continuity of rotation representations in neural networks.
\newblock \emph{2019 IEEE/CVF Conference on Computer Vision and Pattern Recognition (CVPR)}, 2018.

\bibitem[Zhou and Wang(2023)]{Zhou_2023_ude}
Zixiang Zhou and Baoyuan Wang.
\newblock Ude: A unified driving engine for human motion generation.
\newblock In \emph{Proceedings of the IEEE/CVF Conference on Computer Vision and Pattern Recognition (CVPR)}, pages 5632--5641, 2023.

\end{thebibliography}
}


\end{document}



\maketitlesupplementary
In this supplementary,  we first provide a detailed exposition of the implementation specifics of our proposed method. We then describe the data capture process and present additional results from our dataset. Finally, we showcase further qualitative results to demonstrate the effectiveness of our approach.
\section{Implementation Details}
\subsection{Trajectory Generation Model}\label{sec:TGM}
\myparagraph{Network Architecture.} 
We employ a transformer model with 8 encoder-only layers, each comprising 4 attention heads, to learn the mapping from user input to diverse trajectories. The input dimension of the transformer is set to 256 with the feedforward dimension set to 1024 and the dropout rate set to 0.1. Following the \textit{Seperate Condition Tokenization(SCT)} method proposed in CAMDM~\cite{camdm}, we first use separate tokenizers for each input condition and then concatenate the condition tokens with the noisy motion sequence. Specifically, we use linear layers to transform the soccer skill label $\textbf{S}\in\mathbb{R}^{1} \mapsto \mathbb{R}^{256}$, target trajectory point $\textbf{G}\in\mathbb{R}^{8}\mapsto\mathbb{R}^{256}$ and past trajectory $\textbf{T}^{\mathcal{P}}\in\mathbb{R}^{\mathcal{P}\times8}\mapsto\mathbb{R}^{\mathcal{P}\times256}$. After concatenating with the noise, the final input dimension is $\mathbb{R}^{(\mathcal{P}+\mathcal{F}+2)\times256}$. During the forward pass, we add the input with a standard positional embedding. The encoded features are then processed through the transformer encoder and the trajectory output $T \in \mathbb{R}^{\mathcal{F}\times256}\mapsto\mathbb{R}^{\mathcal{F}\times8}$ is obtained via a final linear layer.

\myparagraph{Training.}
In training, the reconstruction loss and velocity loss are specifically defined as follows:
\begin{equation}
    \mathcal{L}_{\text{recon}} = \frac{1}{F} \sum_{i=1}^{F} ||z_{0}^{i} - \hat{z}_{0}^{i}||_{2}^{2},
    \label{eq:recon loss}    
\end{equation}
\begin{equation}
    \mathcal{L}_{\text{vel}} = \frac{1}{F-1} \sum_{i=1}^{F-1} ||(z_{0}^{i+1} - z_{0}^{i}) - (\hat{z}_{0}^{i+1} - \hat{z}_{0}^{i})||_{2}^{2},
    \label{eq:vel loss}
\end{equation}
where $z_{0}$ represents the groundtruth trajectory extracted from the training data and $\hat{z}_{0}$ represents the model predicted future trajectory, superscript $i$ represents the frame index. The two training losses share the same weight $\lambda_{recon}=\lambda_{vel}=1$. During training, we optimize the model using the AdamW optimizer~\cite{Loshchilov2017DecoupledWD} with a weight decay of 0.01 and a learning rate of $10^{-4}$. Training is conducted for 4000 epochs on a single NVIDIA GeForce RTX 4090 GPU, with the batch size set to 512.
\subsection{Soccer Motion Generation Model}
\myparagraph{Network Architecture.} Our soccer motion generation model adopts a transformer-based architecture with 4 encoder-only layers, following a structure similar to the trajectory generation model described in Sec.~\ref{sec:TGM}. We use linear layers to transform the soccer skill label $\textbf{S}\in\mathbb{R}^{1} \mapsto \mathbb{R}^{256}$, past soccer motion $\mathbf{X}^{\mathcal{P}}\in\mathbb{R}^{\mathcal{P}\times(28\times6)}\mapsto\mathbb{R}^{\mathcal{P}\times256}$ and future trajectory $\mathbf{T}^{\mathbb{R}^{\mathcal{F}\times8}}\mapsto\mathbb{R}^{\mathcal{F}\times256}$. Note that here we map the human root position $h_p\in\mathbb{R}^{3}\mapsto\mathbb{R}^{6}$ and ball state $b\in\mathbb{R}^{3+3+1}\mapsto\mathbb{R}^{6+6+6}$ using zero padding. We then concatenate the human root position and ball state with human joint rotation $h_\theta\in\mathbb{R}^{24\times6}$ to get the soccer motion $X\in\mathbb{R}^{28\times6}$, which serves as the input and output to the network.

\myparagraph{Training.}
During training, the loss weights are set to $\lambda_{pos} = \lambda_{vel} = \lambda_{foot} = 1$. The model is optimized using the AdamW optimizer~\cite{Loshchilov2017DecoupledWD} with a weight decay of 0 and a learning rate of $5 \times 10^{-4}$. Training is conducted for 1000 epochs on a single NVIDIA GeForce RTX 4090 GPU, with a batch size of 512.

\begin{figure*}[t!]
    \centering
    \includegraphics[width=\textwidth]{figures/setting.pdf}
    \caption{Motion capturing setup, the blue cameras are OptiTrack Prime x13 cameras for capturing soccer motion.}
    \label{fig:dome setting}
\end{figure*}
\begin{figure*}[t!]
    \centering
    \includegraphics[width=\textwidth]{figures/supplementary_dataset_gallery.pdf}
    \caption{More results of our dataset. We provide detailed descriptions for each action category.}
    \label{fig:dataset gallery}
\end{figure*}

\subsection{Contact Guidance Module}
\myparagraph{Hyperparameters.}
In the contact guidance module, several hyperparameters are introduced, including the acceleration threshold and the penalty term for a lifted foot. This section details the methodology used to determine these hyperparameters. During the data preprocessing stage, ball-foot contacts within the dataset are annotated. Initially, a subset of contact instances is manually labeled. Subsequently, a statistical analysis is conducted to establish distance thresholds for ball-foot contact, as defined by the following equation:
\begin{equation}
    c_{b}^{j} = \mathbb{I} ((b_{p} - f_{p}^{j}) \leq \tau_{d}),
    \label{eq:ball-foot contact}
\end{equation}
where $c_{b}^{j}$ represents the contact between the ball and the joint, superscript $j$ denotes the joint index, $b_{p}$ is the global ball position, $f_{p}^{j}$ is the global foot joint position, and $\tau_{d} = 0.1m$ is the distance threshold.
Based on the distance-based ball-foot contact annotations, we further determine contacts using the ball's acceleration. The accuracy of contact determination is evaluated under various acceleration thresholds $\tau_{a}$, as shown in Tab.~\ref{tab:acceleration thresholds}. A threshold that is too low leads to unintended contact enforcement, causing the foot and ball to stick together continuously. Conversely, an excessively high threshold eliminates meaningful ball-foot contact, rendering this module ineffective. Based on these observations, the acceleration threshold is ultimately set to $2m/s^{2}$.
\begin{table}[t]
\centering
\renewcommand\arraystretch{1.1}
\resizebox{1\linewidth}{!}{
\begin{tabular}{ccccccc}
\toprule[2pt]
\multicolumn{2}{c}{$\tau_{a}$} 
& $0.5m/s^2$ & $1m/s^2$ & $1.5m/s^2$ & $2m/s^2$ & $3m/s^2$ \\
\cline{1-7}
\multicolumn{2}{c}{Acc.$\uparrow$}   
& $56.66\%$ &  $67.13\%$  &  $79.12\%$ & $\mathbf{91.47\%}$ & $83.31\%$ \\ 
\bottomrule[2pt]
\end{tabular}
}
\caption{Accuracy of different acceleration thresholds.}
\label{tab:acceleration thresholds}
\vspace{-1em} 
\end{table}
For the penalty term $w_{d}$, we conduct a statistical analysis on the dataset. The dataset includes annotations indicating which joint contacts the ball, as determined by Eq.~\ref{eq:ball-foot contact}. By applying the acceleration threshold, we calculate the required contact duration for the ball and identify the specific joint responsible for contact based on distance, with accuracy results presented in Tab.~\ref{tab:penalty term}. When the penalty term is set to 1, it causes the ball to stick to the foot when the distance between the foot and the ball is very small. Conversely, an excessively large penalty term prevents the grounded foot from making contact with the ball.  Ultimately, we set the penalty term to 2. The hyperparameters can be fine-tuned based on factors such as the friction of the environment and the size of the ball to ultimately achieve the most effective performance.
\begin{table}[t]
\centering
\renewcommand\arraystretch{1.1}
\resizebox{1\linewidth}{!}{
\begin{tabular}{ccccccc}
\toprule[2pt]
\multicolumn{2}{c}{$w_{d}$} 
& 1 & 1.5 & 2 & 2.5 & 3 \\
\cline{1-7}
\multicolumn{2}{c}{Acc.$\uparrow$}   
& $88.31\%$ &  $89.19\%$  &  $\mathbf{94.43\%}$ & $91.41\%$ & $87.01\%$ \\ 
\bottomrule[2pt]
\end{tabular}
}
\caption{Accuracy of different penalty terms.}
\label{tab:penalty term}
\vspace{-1em} 
\end{table}

\subsection{Implementation Details}
\myparagraph{Runtime Analysis.}
We provide a detailed analysis of the efficiency of each component in our framework. In the trajectory generation stage, the single-step diffusion model deployed in the Unity engine, combined with the blending strategy, takes approximately 8ms. In the soccer motion generation stage, due to the addition of 2 contact guidance steps in the diffusion process, the computation of the loss function and backward pass increases the inference time. Consequently, our 8-step denoising diffusion model takes approximately 80ms in total. Lastly, the network transmission time from Python to Unity is about 7ms. Overall, our method takes approximately 95ms for the whole inference process. Future work could explore reducing the computation time for contact guidance to further optimize the efficiency of the algorithm.

\begin{table}[t]
\centering
\renewcommand{\arraystretch}{1.5}
\begin{tabular}{|c|c|c|}
\hline
\textbf{Category}      & \textbf{Property}       & \textbf{Value}        \\
\hline
\multirow{9}{*}{\textbf{Football}} 
                       & Mass                   & 0.43 kg               \\
\cline{2-3}
                       & Drag                   & 0.2                   \\
\cline{2-3}
                       & Angular Drag           & 0.05                  \\
\cline{2-3}
                       & Radius                 & 0.11 m                \\
\cline{2-3}
                       & Dynamic Friction       & 0.5                   \\
\cline{2-3}
                       & Static Friction        & 0.5                   \\
\cline{2-3}
                       & Bounciness             & 0.1                   \\
\cline{2-3}
                       & Friction Combine       & Multiply              \\
\cline{2-3}
                       & Bounce Combine         & Maximum               \\
\hline
\multirow{5}{*}{\textbf{Ground}} 
                       & Dynamic Friction       & 0.6                   \\
\cline{2-3}
                       & Static Friction        & 0.6                   \\
\cline{2-3}
                       & Bounciness             & 0.05                  \\
\cline{2-3}
                       & Friction Combine       & Average               \\
\cline{2-3}
                       & Bounce Combine         & Minimum               \\
\hline
\multirow{2}{*}{\textbf{Environment}} 
                       & Gravity                & -9.81 m/s² (Y-axis)   \\
\cline{2-3}
                       & Fixed Timestep         & 0.01 s                \\
\hline
\end{tabular}
\caption{Physics settings for simulating soccer ball movement in Unity.}
\label{tab:unity}
\end{table}

\section{Dataset Details}
\myparagraph{Capture setting.}
We provide the capture details of the Soccer-X dataset. During motion capturing, players wore motion capture suits equipped with 41 markers, leveraging the baseline skeletal framework provided by Motive software~\cite{Optitrack}. The soccer ball was asymmetrically marked with 12 markers to ensure accurate tracking and was defined as a rigid body.
Our motion capture setup consists of 16 cameras arranged in a surrounding configuration. Fig.~\ref{fig:dome setting} shows the motion capture setup, where the blue cameras indicate those used during the capture process.

The dataset includes annotations for action name and ball-foot contact events. As shown in Fig.~\ref{fig:dataset gallery}, we also provide a detailed description of each action category.

\myparagraph{Shooting simulation.}
For the shooting motion, due to field constraints, we were only able to record the trajectory data of the soccer ball during the first half of its movement. To address this limitation, we extracted the position and rotation information from the last two frames of the recorded data and calculated the linear velocity and angular velocity of the soccer ball in the final frame. Using these initial conditions, along with other necessary physical parameters, we simulated the second half of the trajectory in the Unity engine. The specific parameter settings are shown in Tab.~\ref{tab:unity}.

In this context, Football is a GameObject with a Rigidbody component and a Sphere Collider component, Ground is a GameObject with a Collider component, and Environment refers to Unity's environment settings. The mass and radius of the football were set strictly according to the standards of the Fédération Internationale de Football Association (FIFA) and were consistent with the actual mass and dimensions of the soccer ball used in the experiment. Other parameters (such as friction coefficients and elasticity coefficients) were fine-tuned through multiple tests to ensure the simulation results closely matched with the real-world motion behavior.

\section{More Experiment}
\paragraph{More Results.}
We further validate the diversity of the soccer motions generated by our method. As shown in Fig.~\ref{fig:result diversity}, under identical user control input(including the same character speed, direction, and soccer skill label), our method generates soccer motions that exhibit different dribbling styles. This demonstrates that our trajectory generation model and soccer motion generation model have learned diverse character trajectories and soccer skills from our training data.

\paragraph{More Comparisons.}
We further compare our method with baseline approaches. Specifically, we use a complete sequence of the step-over skill in the trick category for a clearer comparison. As shown in Fig.~\ref{fig:comparison_supp}, the soccer motions generated by LMP~\cite{10.1145/3386569.3392450} and MANN-DP~\cite{mann,deepphase} exhibits unnatural ball-foot contacts, while CM~\cite{codebookmatching} fails to generate the corresponding step-over skill. In contrast, our method produces realistic and natural step-over motions.

\begin{figure*}[t!]
    \centering
    \includegraphics[width=\textwidth]{figures/diversity.pdf}
    \caption{Additional qualitative results of SMGDiff. Given the same user control input, SMGDiff generates diver soccer motions.}
    \label{fig:result diversity}
\end{figure*}

\begin{figure*}[t!]
    \centering
    \includegraphics[width=\textwidth]{figures/Comparison_supp.pdf}
    \caption{Additional qualitative comparisons. SMGDiff outperforms baselines on the step-over trick.}
    \label{fig:comparison_supp}
\end{figure*}

{
    \small
    \bibliographystyle{ieeenat_fullname}
    \bibliography{main}
}
